\documentclass{article}

\usepackage[preprint]{neurips_2020}
\usepackage[utf8]{inputenc} % allow utf-8 input
\usepackage[T1]{fontenc}    % use 8-bit T1 fonts
\usepackage{hyperref}       % hyperlinks
\usepackage{url}            % simple URL typesetting
\usepackage{booktabs}       % professional-quality tables
\usepackage{amsfonts}       % blackboard math symbols
\usepackage{nicefrac}       % compact symbols for 1/2, etc.
\usepackage{microtype}      % microtypography
\usepackage{graphicx}
\usepackage{amsmath}
\usepackage{amsthm}
\usepackage{colortbl}
\usepackage{times}
\usepackage{soul}
\usepackage{booktabs}
\usepackage{algorithm}
\usepackage{algorithmic}
\usepackage{comment}
\usepackage{pifont}
\usepackage{soul}
\usepackage{enumitem}

\usepackage[hang,flushmargin]{footmisc}
\usepackage{subfig}
\usepackage{wrapfig}
\usepackage[font=scriptsize,labelfont=bf]{caption}
\usepackage[belowskip=-5pt,aboveskip=3pt]{caption} % re-arrange tables on last page, select
\usepackage{sidecap}
\usepackage{hyperref}
\hypersetup{
    colorlinks=true,
    linkcolor=blue,
    filecolor=magenta,      
    urlcolor=magenta, %cyan,
    citecolor=blue,
}

\def\argmin{\operatornamewithlimits{arg\,min}}
\def\argmax{\operatornamewithlimits{arg\,max}}

\definecolor{tabgray1}{rgb}{0.95,0.95,0.95}
\definecolor{tabgray2}{rgb}{0.85,0.85,0.85}
\definecolor{top1}{rgb}{0.99, 0.76, 0.8}%{1.0, 0.6, 0.6} %{0.96, 0.94, 0.93} \definecolor{cerise}{rgb}{0.87, 0.19, 0.39}
\definecolor{top2}{rgb}{0.94, 0.9, 0.55}
\definecolor{top3}{rgb}{0.67, 0.9, 0.93} 

\usepackage{color}
\definecolor{mygray}{gray}{0.6}
\definecolor{tabgray2}{rgb}{0.85,0.85,0.85}
\definecolor{tabgray3}{rgb}{0.8,0.8,0.8}
\definecolor{dg}{rgb}{0,0.694,0.298}
\definecolor{purple}{rgb}{0.4,0.176,0.569}
\definecolor{pink}{cmyk}{0, 0.7808, 0.4429, 0.1412}
\newcommand{\fei}[1]{\textbf{\textcolor{dg}{Fei: #1}}}

\newcommand{\felix}[1]{\textbf{\textcolor{magenta}{Felix: #1}}}
\newcommand{\qing}[1]{\textbf{\textcolor{pink}{Qing: #1}}}
\newcommand{\first}[1]{\textcolor{top1}{#1} }
\newcommand{\second}[1]{\textcolor{top2}{#1} }
\newcommand{\third}[1]{\textcolor{top3}{#1} }

\usepackage{xspace}
\makeatletter
\DeclareRobustCommand\onedot{\futurelet\@let@token\@onedot}
\def\@onedot{\ifx\@let@token.\else.\null\fi\xspace}
\def\eg{\emph{e.g}\onedot} 
\def\ie{\emph{i.e}\onedot} 
\def\cf{\emph{c.f}\onedot} 
\def\etc{\emph{etc}\onedot} 
\def\wrt{w.r.t\onedot} 

\makeatother

\title{Watch out! Motion is \textcolor{mygray}{Blurring}$\kern-60pt${Blurring} the Vision of Your Deep Neural Networks}

\author{
  Qing Guo$^1$ \And Felix Juefei-Xu$^2$ \And Xiaofei Xie$^{1*}$ \And Lei Ma$^3$\thanks{Xiaofei Xie and Lei Ma are corresponding authors~(\href{mailto:xfxie@ntu.edu.sg}{xfxie@ntu.edu.sg}, \href{mailto:malei@ait.kyushu-u.ac.jp}{malei@ait.kyushu-u.ac.jp}).} \AND Jian Wang$^1$ \And Bing Yu$^3$ \And Wei Feng$^4$ \And Yang Liu$^1$ 
    \AND 
    \normalfont
  $^1$Nanyang Technological University, Singapore \quad $^2$Alibaba Group, USA \\
  $^3$Kyushu University, Japan \quad $^4$Tianjin University, China \\
}

\begin{document}

\maketitle

% \vspace{-1.5em}
\begin{abstract}
% \vspace{-0.9em}
The state-of-the-art deep neural networks (DNNs) are vulnerable to adversarial examples with additive random noise-like perturbations.
While such examples are hardly found in the physical world, the image blurring effect caused by object motion, on the other hand, commonly occurs in practice, making the study of which greatly important especially for the widely adopted real-time image processing tasks (\eg, object detection, tracking).
In this paper, we initiate the first step to comprehensively investigate the potential hazards of blur effect for DNN, caused by object motion.
We propose a novel adversarial attack method
that can generate visually natural motion-blurred adversarial examples, named motion-based adversarial blur attack (AB$\kern-3pt$\textcolor{mygray}{B}A).
To this end, we first formulate the kernel-prediction-based attack where an input image is convolved with kernels in a pixel-wise way, and the misclassification capability is achieved by tuning the kernel weights.
To generate visually more natural and plausible examples, we further propose the saliency-regularized adversarial kernel prediction, where the salient region serves as a moving object, and the predicted kernel is regularized to achieve visual effects that are natural. Besides, the attack is further enhanced by adaptively tuning the translations of object and background.
A comprehensive evaluation on the NeurIPS'17 adversarial competition dataset demonstrates the effectiveness of AB$\kern-3pt$\textcolor{mygray}{B}A by considering various kernel sizes, translations, and regions. 
The in-depth study further confirms that our method shows more effective penetrating capability to the state-of-the-art GAN-based deblurring mechanisms compared with other blurring methods. We release the code to \url{https://github.com/tsingqguo/ABBA}.
% to our methods.
% Furthermore, we study the effects of state-of-the-art GAN-based deblurring mechanisms to our methods.
\end{abstract}

%----------------------------------------------------------------------
%----------------------------------------------------------------------
\vspace{-1.2em}
\section{Introduction}\vspace{-1em}

% \felix{emphasize on the pervasiveness of the problem, long exposure, low-light, very common in real world application, emphasize on the first work in this direction. xxx is lacking.... potentially important vulnerability in real world. pressing problem... no need to mention too much background. Blur is a main aspect of photography and computational photography, found in real-world. Identifying and improve the resilience of xxx is}

Deep neural networks (DNN) have been widely applied in various vision perception tasks (\eg, object recognition, segmentation, scene understanding), permeating many aspects of our daily life, such as autonomous driving, robotics, video surveillance, photo taking, \etc. However, the state-of-the-art DNNs are still vulnerable to adversarial examples. Extensive previous works are proposed (\eg, FGSM \cite{GoodfellowARXIV2014}, BIM \cite{KurakinICLR2017}, MI-FGSM \cite{DongCVPR2018}, C\&W \cite{CW_2017_SSP}), to mislead the DNN through additive noise perturbations that could be obtained by optimizing the adversarial objectives. 
To be imperceptible to human, $L_\mathrm{p}$-norm plays an important role in such attacks, confining the perturbation noise to be small.
However, the random noise-like perturbation 
often does not pose imminent threats to the camera systems, which does not usually occur in natural environment.
% current state-of-the-art adversarial noises can only be obtained theoretically and rarely in natural environment, which does not yet pose imminent threats to the camera systems.
Thus, some recent attempts \cite{KurakinICLR2017} were made to physically fashion adversarial examples such as by putting up stickers or printed patterns on the physical stop sign, \etc. Again, these artifacts are often intentionally prepared  
for adversarial attacks, which are not `naturally' found in the real-world environment either.

While the blurring effect caused by object motion commonly occurs in practical image perception systems, the potential hazards of motion blur effect to the DNN are largely untouched so far. 
Motion blur naturally happens during the exposure time of image capturing.
When an object moves at a relatively high speed, all information of the object during the image capture process is integrated, constituting a blur-like image along a relative moving direction. Compared with other kinds of image blur (\eg, defocus blur caused by using unsuitable camera focus), motion blur is directly related to the motion of object and camera, whose effect cannot be easily removed by adjusting the camera's setting. As a result, motion blur almost coexists with the camera and potentially posts serious effects on DNN perception-based systems. However, up to present, there are limited studies discuss how motion blur affects the DNN perception tasks. It is not even clear whether and what kinds of motion blur can systematically mislead a DNN.

\begin{wrapfigure}{r}{0.43\columnwidth}
\vspace{-1em}
% \begin{figure}[ht]
% \centering
% \includegraphics[width=0.7\linewidth]{./fig1.eps}
\includegraphics[width=0.43\columnwidth]{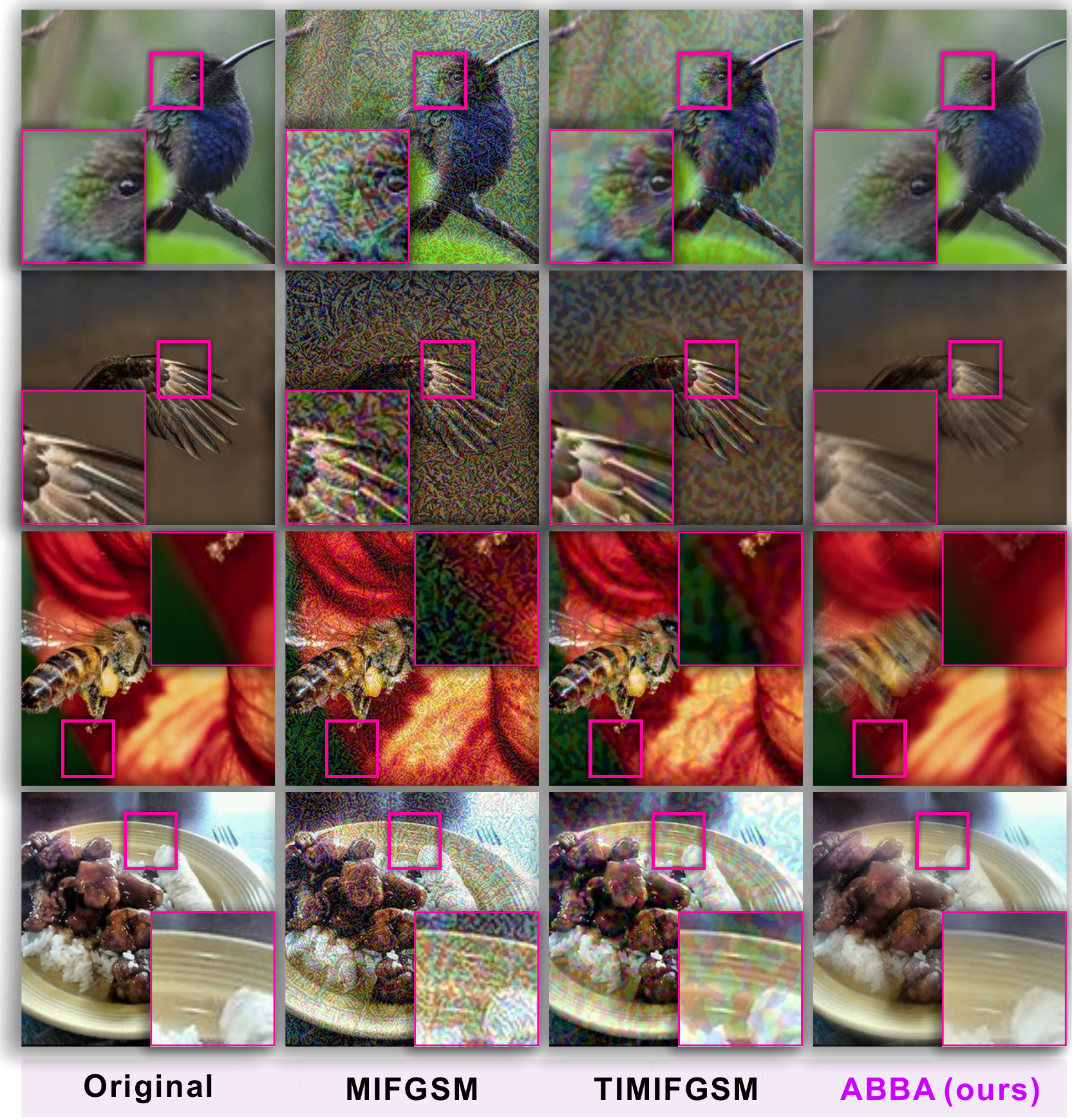}
\caption{Four adversarial examples of MIFGSM \cite{DongCVPR2018}, TIMIFGSM \cite{DongCVPR2019} and our AB$\kern-3pt$\textcolor{mygray}{B}A. MIFGSM and TIMIFGSM produces apparent noise on all four cases. Our AB$\kern-3pt$\textcolor{mygray}{B}A generates visually natural motion blur. All adversarial images fool the Inception-v3 model.}
\label{fig:motivation}
% \end{figure}
\vspace{-1em}
\end{wrapfigure}

In this paper, we initiate the first step to comprehensively investigate the blur effects to DNNs from the adversarial attack perspective, where systematic motion blur-based adversarial example discovery would be an important step towards further DNN enhancement.
In particular, we propose a new type of adversarial attack, termed motion-based adversarial blur attack (AB$\kern-3pt$\textcolor{mygray}{B}A), which can generate visually natural and plausible motion-blurred adversarial examples. 
We first formulate the kernel-prediction-based attack where an input image is convolved with kernels in a pixel-wise way, and the misclassification ability is guided by systematically tuning the kernel weight. In order to generate more natural motion blurred examples, we also propose the saliency-regularized adversarial kernel prediction, where the salient region serves as a moving object, and the predicted kernel is regularized to achieve visually natural effects. Besides, our method could easily adjust the blur effects of different exposure time during the image capturing in the real world, by adaptively tuning translations of object and background.
%the attack is further enhanced by adaptively tuning translations of object and background.
% thus leads to visually realistic adversarial examples.
%
% Such saliency-regularized motion blur happens only locally and consistently. The overall image is outstandingly visually realistic and indistinguishable from an actual moving object.

% Although it can have a larger perturbation magnitude within the bound of an object, due to the blur effect, the overall image is outstandingly visually realistic and indistinguishable from an actual moving object captured by a slightly longer exposure, a scenario that can actually occur in the real world.

We perform comprehensive evaluation on the effectiveness of the proposed AB$\kern-3pt$\textcolor{mygray}{B}A, benchmarked against various noise-based attacks on both attack success rates and transferability. The main contributions of this work can be summarized as follows. \ding{182} To the best of our knowledge, we make the very first attempt to investigate kernel-based adversarial attack. \ding{183} We propose a motion-based adversarial blur attack as a new type of attack mode, to be added to the adversarial attack family. \ding{184} In order to produce more visually plausible blur attack, we introduce a saliency regularizer that forces consistent blur patterns within the boundary of the objects (or background in some cases). \ding{185} Compared with the state-of-the-art (SOTA)
additive-noise based adversarial attacks and common blur techniques, our proposed method achieves better attack success rate and transferability. \ding{186} Furthermore, our proposed method has demonstrated higher penetration capability against the SOTA GAN-based deblur mechanism, compared to normal image motion blur.

%----------------------------------------------------------------------
%----------------------------------------------------------------------
% \section{Related Work}\vspace{-1em}
% \label{sec:related_work}

\textbf{Related Work.} Since the discovery of adversarial examples to attack a deep neural network (DNN) both theoretically \cite{GoodfellowARXIV2014} and physically \cite{KurakinICLR2017}, there has been extensive research towards developing adversarial attack and defense mechanisms. The basic iterative method (BIM) \cite{KurakinICLR2017}, the C\&W method \cite{CW_2017_SSP}, the fast gradient signed method (FGSM) \cite{GoodfellowARXIV2014}, and the momentum iterative fast gradient sign method (MI-FGSM) \cite{DongCVPR2018}, \etc, are a few popular ones among early adopters in the research community. Building upon these ideas, researchers have been continuously pushing the envelope in many ways. For example, serious attempts have been made to integrate momentum term into the iterative process for the attacks \cite{DongCVPR2018}. By doing so, the momentum can help stabilize the update directions, begetting more transferable adversarial examples and posing more threats to adversarially trained defense mechanisms. More recently, \cite{DongCVPR2019} proposes to optimize the noise perturbation over an ensemble of translated images, making the generated adversarial examples more robust against white-box models being attacked while achieving better transferability. 
The mainstream adversarial attack is an additive noise pattern that is learnable given the model parameters under a white-box setting. Perhaps the prevalence is partially due to the fact that the adversarial noise with the `\emph{addition}' operation is relatively straightforward to optimize for. Of course, there are many other ways to alter a benign image beyond the addition operation that are all potential candidates for coming up with new types of adversarial attack modes. One caveat of additive noise attack is the lack of balance between being visually plausible and imperceptible while having high attack success rate. Usually, it has to compromise one for the other. Researchers are looking beyond additive noise attack to seek novel attack modes that strike a better balance between visual plausibility and performance, \eg, coverage-guided fuzzing~\cite{issta19_deephunter}, multiplicative attack \cite{YeArxiv2019}, deformation attack \cite{XiaoICLR2018,AlaifariICLR2019,wang2020amora}, and semantic manipulation attack \cite{BhattadICLR2020}.

We are proposing a new type of motion-based adversarial blur attack that can generate visually natural and plausible motion-blurred adversarial examples, inspired by kernel prediction  \cite{MildenhallCVPR2018,NiklausCVPR2017,NiklausICCV2017} and motion blur generation \cite{BrooksCVPR2019}. One desired property of the proposed method is the immunity and robustness against the SOTA deblurring techniques (\eg, \cite{KupynCVPR2018,KupynICCV2019}). In an effort to understand black-box DNN better, an image saliency paradigm is proposed \cite{FongICCV2017} to learn where an algorithm looks at, by discovering which parts of an image most affect output score when perturbed in terms of Gaussian blur, replacing the region with constant value and injecting noise. The localization of the blur region is performed through adaptive iteration whilst ours is saliency regularized that leads to a visually plausible motion-blurred image. The major difference is that their Gaussian blur kernel is fixed while ours is learnable to maximally jeopardize the image recognition DNNs.

%----------------------------------------------------------------------
%----------------------------------------------------------------------
\vspace{-0.9em}
\section{Methodology}\vspace{-0.7em}
%
\if 0
In this section, we first revisit the conventional additive-perturbation-based attack and then propose a different attack, called kernel-prediction-based attack. At last, we introduce the motion-based adversarial blur attack, which is an advanced kernel-prediction-based attack. 
\fi

\if 0 \fei{Qing, please confirm my revise} \fi

%----------------------------------------------------------------------
\subsection{Background: Additive-Perturbation-Based Attack}\vspace{-0.5em}
Let $\mathbf{X}^{\text{real}}$ be a real (untampered) example, \eg, images in the ImageNet dataset, and $y$ denotes its ground truth label. A classifier denoted as $\mathrm{f}(\mathbf{X}):\mathcal{X}\to\mathcal{Y}$ predicts the label of $\mathbf{X}^{\text{real}}$. An attack method aims to generate an adversarial example denoted as $\mathbf{X}^{\text{adv}}$ that can fool the classifier to predict an incorrect label with imperceptible perturbation. Existing attack methods mainly focus on the additive adversarial perturbation that is added to the real example to get $\mathbf{X}^\text{adv}=\mathrm{g}(\mathbf{X}^{\text{real}},\delta)=\mathbf{X}^{\text{real}}+\delta$, 
where $\delta$ is generated by maximizing a loss function $J(\mathbf{X}^{\text{adv}},y)$ with a constrained term: 
\begin{align}\label{eq:add_adv_obj}
\argmax_{\delta} J(\mathbf{X}^{\text{real}}+\delta,y) \mathrm{~~subject~to~~} \|\delta\|_{\mathrm{p}}\leq\epsilon_\mathrm{a},
\end{align}
% \end{wrapfigure}
%---------------------------
% where $\|\cdot\|_\mathrm{p}$ is the $L_{\mathrm{p}}$ norm.
% and the $L_{\infty}$ norm is widely used. 
% As an optimization problem, 
For example, gradient descent is widely employed by many methods to generate adversarial examples, \eg, FGSM, BIM, MIFGSM, DIM, and TIMIFGSM.
% to solve the problem and inspire several successful attack methods, \eg, FGSM, BIM, MIFGSM, DIM, TIMIFGSM, \etc.

%----------------------------------------------------------------------
\vspace{-0.7em}
\subsection{AB$\kern-3pt$\textcolor{mygray}{B}A$_\text{pixel}$: Kernel-Prediction-Based Adversarial Attack}\vspace{-0.5em}
\label{subsec:kernel-prediction}
Besides `+', there are various techniques that can perform advanced image transformation for different objectives, \eg, Gaussian filter for image denoising, Laplacian filter for image sharpening, and guided filter for edge-preserving smoothing \cite{HeTPAMI2013}, which are all kernel-based techniques, processing each pixel of the image with a hand-crafted or guided kernel. In general, compared with the addition, kernel-based operation can handle more complex image processing tasks via different kinds of kernels.
More recently, several works \cite{BakoACMTOG2017,NiklausCVPR2017,NiklausICCV2017} have found that the kernel weights can be carefully predicted for advanced tasks with high performance, \eg, high quality noise-free rendering and video frame interpolation. Inspired by these works, in this paper, we propose the kernel-prediction-based attack. Specifically, we process each pixel (\ie, $\mathbf{X}^{\text{real}}_p$) of a real example $\mathbf{X}^{\text{real}}$ with a kernel $\mathbf{k}_p$,
% \eg, $\mathbf{X}^{\text{real}}_p$,
% within 
%---------------------------
\begin{align}\label{eq:kernel_adv}
\mathbf{X}^\text{adv}_p=\mathrm{g}(\mathbf{X}^{\text{real}}_p,\mathbf{k}_p,\mathcal{N}(p))=\sum_{q\in\mathcal{N}(p)}\mathbf{X}^{\text{real}}_qk_{pq},
\end{align}
%---------------------------
where $p$ denotes the $p$-th pixel in $\mathbf{X}^\text{adv}$ and $\mathbf{X}^\text{real}$, \ie, $\mathbf{X}^\text{adv}_p$ and $\mathbf{X}^{\text{real}}_p$, respectively. $\mathcal{N}(p)$ is a set of $N$ pixels in $\mathbf{X}^\text{real}$ and $p\in\mathcal{N}(p)$. The kernel $\mathbf{k}_p$ also has the size of $N$ and determines the weights of $N$ pixels in $\mathcal{N}(p)$. In general, we have $\sum_{q\in\mathcal{N}(p)}k_{pq}=1$ to ensure the generated image lies within the respective neighborhood of the input image, where a softmax activation is often adopted for this requirement \cite{BakoACMTOG2017}. 
To better understand Eq.~(\ref{eq:kernel_adv}), we discuss with two simple cases: 1) when we let $\mathcal{N}(p)$ be a neighborhood of the pixel $p$ and the kernel of each pixel is a fixed Gaussian kernel, $\mathbf{X}^{\text{adv}}$ is the Gaussian-blurred $\mathbf{X}^{\text{real}}$. Similarly, we can obtain defocus-blurred image with disk kernels. 2) when we set $\mathrm{max}(\mathbf{k}_{p})=k_{pp}$ and $\forall q,\,k_{pq}\neq0$, the perturbation of $\mathbf{X}^{\text{adv}}$ becomes more imperceptible as length of $\mathbf{k}_p$ decreases. 
%
% $\bigcup_{\forall p\in \mathbf{X}^\text{real}} F_{i}$
%
% \bigcup_{\forall p\in \mathbf{X}^\text{real}}\{\mathbf{k}_p\}$
%$\mathcal{K}=\{\forall p\in \mathbf{X}^\text{real}|\mathbf{k}_p\}$,
%
To achieve high attack success rate, we need to optimize kernels of all pixels independently, \ie, $\mathcal{K}=\{\mathbf{k}_p|\forall p~\text{in}~\mathbf{X}^\text{real}\}$, according to the loss function of AI-related tasks, \eg, image classification, and constrained terms
%---------------------------
% \begin{align}\label{eq:kernel_adv_obj}
% \argmax_{\mathcal{K}}  J \Bigg( \Big\{\sum_{q\in\mathcal{N}(p)}\mathbf{X}^{\text{real}}_qk_{pq}\Big\},y\Bigg) ~~~~~~~\\
% \mathrm{~~~~subject~to~~~~}  \forall p, \|\mathbf{k}_p\|_{0}\leq\epsilon, \max(\mathbf{k}_p)=k_{pp}.\nonumber
% \end{align}
%---------------------------
%---------------------------
% \begin{align}\label{eq:kernel_adv_obj}
% \argmax_{\mathcal{K}}  J(\{\sum_{q\in\mathcal{N}(p)}\mathbf{X}^{\text{real}}_qk_{pq}\},y)
% ~~~~~~~~~~~~~~~~~~~~~\\
% \mathrm{~~subject~to~~} \forall p, \|\mathbf{k}_p\|_{0}\leq\epsilon, \max(\mathbf{k}_p)=k_{pp},\sum_{q\in\mathcal{N}(p)}k_{pq}=1, \nonumber
% \end{align}
%---------------------------
%---------------------------
\footnotesize
\begin{align}\label{eq:kernel_adv_obj}
\argmax_{\mathcal{K}}  J(\{\sum_{q\in\mathcal{N}(p)}\mathbf{X}^{\text{real}}_qk_{pq}\},y)
\mathrm{~~subject~to~~} \forall p, \|\mathbf{k}_p\|_{0}\leq\epsilon, \max(\mathbf{k}_p)=k_{pp},\sum_{q\in\mathcal{N}(p)}k_{pq}=1, 
\end{align}
%---------------------------
\normalsize
%---------------------------
where $\|\mathbf{k}_p\|_{0}$ represents the number of valid kernel elements~(\ie, $\{k_{pq}\neq0\}$) and $\epsilon\in[1,N]$ controls the upper bound of $\|\mathbf{k}_p\|_{0}$. When $\epsilon=1$, we have $\mathbf{X}^\text{adv}=\mathbf{X}^\text{real}$, and when $\epsilon=N$, the perturbation would be the most serious case. 
We can calculate the gradient of the loss function with respect to all kernels, to realize the gradient-based attack. As a result, the attack method can be integrated into any gradient-based additive-perturbation attack methods, \eg, FGSM, BIM, MIFGSM. %In this paper, we adopt the MIFGSM for the high transferability across models which is a significant advantage for real-world security application.
%---------------------------
\begin{wrapfigure}{r}{0.45\columnwidth}
\vspace{-1em}
\includegraphics[width=0.45\columnwidth]{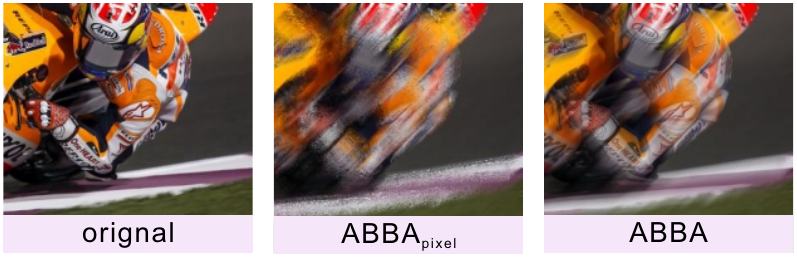}
\caption{From left to right: original image, adversarial examples generated by the kernel-prediction-based attack (AB$\kern-3pt$\textcolor{mygray}{B}A$_\text{pixel}$), and motion-based adversarial blur attack (AB$\kern-3pt$\textcolor{mygray}{B}A).}
\label{fig:kpn}
% \end{figure}
\vspace{-1em}
\end{wrapfigure}
%---------------------------
%
Since the kernel-prediction-based adversarial attack tunes each pixel's kernel independently, it can achieve a significantly high attack success rate, but generating unnatural images that are easily perceptible. For an easier reference, we name this method AB$\kern-3pt$\textcolor{mygray}{B}A$_\text{pixel}$. As shown in Fig.~\ref{fig:kpn}, AB$\kern-3pt$\textcolor{mygray}{B}A$_\text{pixel}$ distorts the original inputs and produces additive noise-like results.
To reach the balance between high attack success rate and natural visual effect, we propose to regularize the kernels to produce visually natural motion blur via the guidance of a visual saliency map. 

%----------------------------------------------------------------------
\vspace{-0.7em}
\subsection{AB$\kern-3pt$\textcolor{mygray}{B}A: Motion-Based Adversarial Blur Attack}%\vspace{-0.5em}
%
% Motion blur is a frequently occurring effect during image capture. We first introduce how to generate visually natural motion-blurred adversarial examples by regularizing the kernel-prediction-based attack. Then, we describe the workflow of our proposed attack in Fig.~\ref{fig:method} for better understanding.

\iffalse
\qing{Please help check following description.}
Note, attacking an image via motion blur allow us to produce visually natural adversarial example with large pixel variations.
%
From the view of conventional additive-perturbation-based attack, our method cannot generate imperceptible adversarial examples, since we can easily perceive apparent change when comparing our results with the original one. However, when seeing our adversarial example alone, \eg, Fig.~\ref{fig:kp_vs_abba}, we can hardly decide if it is a real blurred image or an adversarial example \felix{\ie, the blur operation itself is very realistic.}. 
\fi

\vspace{-0.7em}
\subsubsection{Saliency-Regularized Adversarial Kernel Prediction}\vspace{-0.5em}
%

%-----------------------
% Felix: side caption figure
\begin{SCfigure}
% \begin{figure}[t]
% \vspace{-1em}
    % \centering
    \includegraphics[width=0.67\columnwidth]{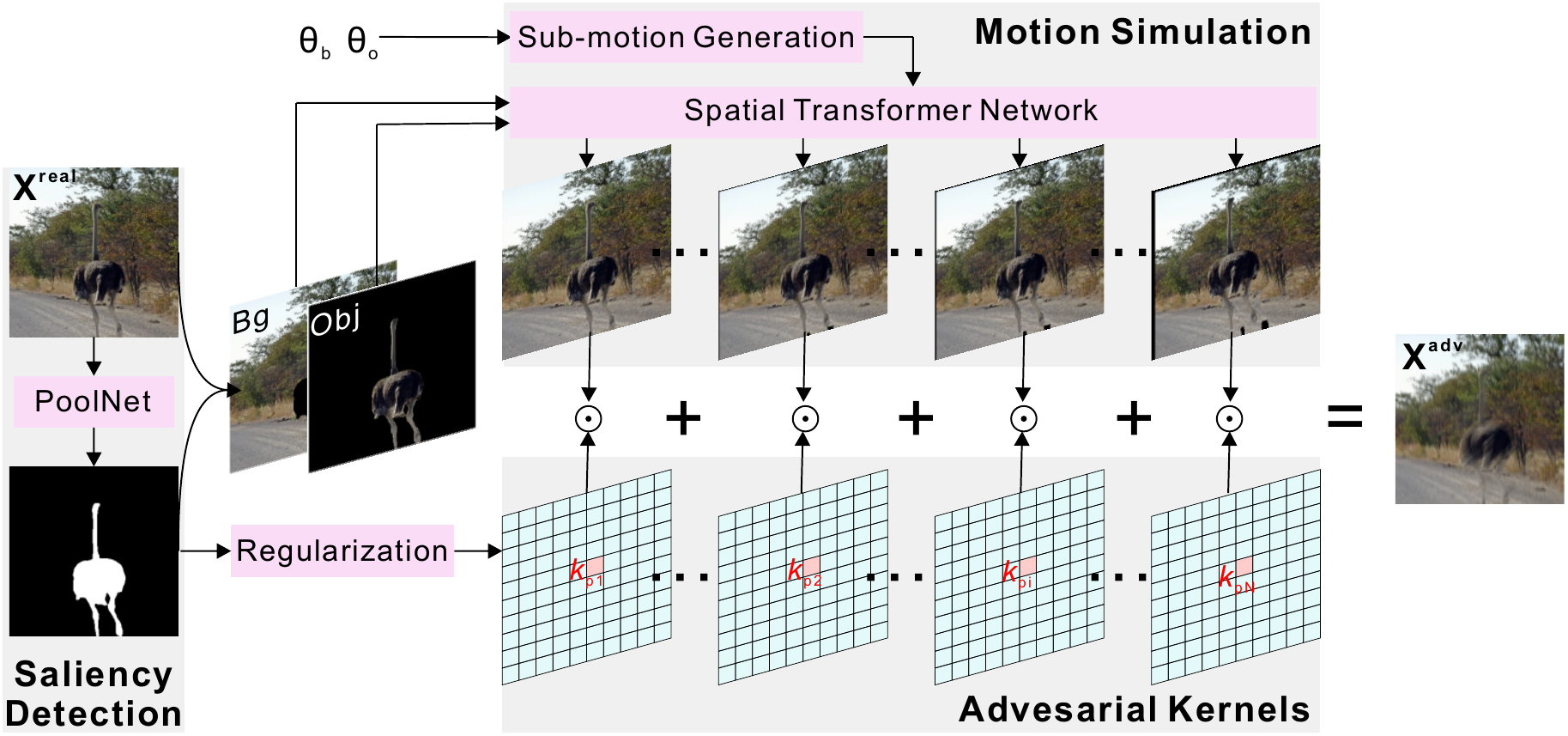}
    \caption{Pipeline of our motion-based adversarial blur attack, \ie, Eq.~(\ref{eq:motion_adv}). First, We use the PoolNet \cite{Liu2019PoolSal} to extract the salient object in the image and obtain the object and background regions. Then, the translation parameters $\theta_{\mathrm{o}}$ and $\theta_{\mathrm{b}}$ are divided to $N$ parts to simulate the motion process and generate $N$ images with the spatial transformer network \cite{JaderbergNIPS2015}. Finally, we get the adversarial example by adding these images with adversarial kernels as weights for each pixel. The adversarial kernels and translation parameters are tuned to realize effective attack by optimizing Eq.~(\ref{eq:motion_adv_obj}). $\odot$ denotes the element-wise product.}
    % \vspace{-1.5em}
\label{fig:method}
% \end{figure}
\end{SCfigure}
%-----------------------

%
Motion blur is often generated during the exposure time by integrating the light from a moving object. To synthesize the motion blur, we need to know where the object is and specify how it moves. 
To this end, given an image, we first use the SOTA saliency detection method, \ie, PoolNet \cite{Liu2019PoolSal}, to extract the salient object $\mathbf{S}$ from $\mathbf{X}^{\text{real}}$ and assume it is moving at the time when capturing the image. The saliency map $\mathbf{S}$ is a binary image, indicating the salient object region~(\ie, $\mathbf{X}^{\text{real}}\odot\mathbf{S}$ ) and background region~(\ie, $\mathbf{X}^{\text{real}}\odot(1-\mathbf{S})$), as shown in Fig.~\ref{fig:method}.
Then, we specify translation transformations to the object and background, respectively. We denote them as $\mathrm{T}(\mathbf{X}^{\text{real}}\odot\mathbf{S},\theta_\mathrm{o})$ and $\mathrm{T}(\mathbf{X}^{\text{real}}\odot(1-\mathbf{S}),\theta_\mathrm{b})$ that are simplified as $\mathbf{X}^{S,\theta_\mathrm{o}}$ and $\mathbf{X}^{1-S,\theta_\mathrm{b}}$, where $\theta_\mathrm{o}$ and $\theta_\mathrm{b}$ are the translation parameters\footnote{\tiny $\theta_\mathrm{o}$ and $\theta_\mathrm{b}$ are 2D vectors in the range of [0,1] and represent the rates of x-axis and y-axis shifting distance of the object\&background regions \wrt image size.} for the object and background, respectively. 

Since motion blur is the integration of all light during the object moving process, we divide the motion represented by $\theta_\mathrm{o}$ and $\theta_\mathrm{b}$ into $N$ sub-motions~(corresponding to the kernel size $N$ in Eq.~(\ref{eq:kernel_adv})) to simulate blur generation. The sub-motions are represented by $\{i\Delta\theta_\mathrm{o},|i\in[1,N]\}$ and $\{i\Delta\theta_\mathrm{b},|i\in[1,N]\}$, where $\Delta\theta_\mathrm{o}=\theta_\mathrm{o}/N$ and $\Delta\theta_\mathrm{b}=\theta_\mathrm{b}/N$.
Then, we redefine Eq.~(\ref{eq:kernel_adv}) as
%---------------------------
% {\footnotesize
% \begin{align}\label{eq:motion_adv}
% \mathbf{X}^\text{adv}_p =\mathrm{g}(\mathbf{X}^{\text{real}}_p,\mathbf{S},\mathbf{k}_p,\mathcal{N}(p))=\sum_{\begin{subarray}~q=\mathcal{N}(p,i)\\i\in~[1,N]\end{subarray}}
% (\mathbf{X}^{S,i\Delta\theta_\mathrm{o}}_q+\mathbf{X}^{1-S,i\Delta\theta_\mathrm{b}}_q)k_{pq},
% \end{align}
% }
%---------------------------
\footnotesize
\begin{align}\label{eq:motion_adv}
\mathbf{X}^\text{adv}_p =\mathrm{g}(\mathbf{X}^{\text{real}}_p,\mathbf{S},\mathbf{k}_p,\mathcal{N}(p))=\sum_{q=\mathcal{N}(p,i),i\in[1,N]}
(\mathbf{X}^{S,i\Delta\theta_\mathrm{o}}_q+\mathbf{X}^{1-S,i\Delta\theta_\mathrm{b}}_q)k_{pq},
\end{align}
%---------------------------
\normalsize
%---------------------------
where $\mathcal{N}(p)$ is a set of the $p$-th pixel in all translated examples. $\mathbf{X}^{S,i\Delta\theta_\mathrm{o}}$ and $\mathbf{X}^{S,i\Delta\theta_\mathrm{o}}$ denote the object and background images translated $i\Delta\theta_\mathrm{o}$ pixels.
Compared with the attack in Sec.~\ref{subsec:kernel-prediction}, the perturbation amplitude is affected by the kernel and translation parameters. The objective function is defined as
%---------------------------
% {\footnotesize
% \begin{align}\label{eq:motion_adv_obj}
% \argmax_{\mathcal{K},\theta_\mathrm{o},\theta_\mathrm{b}}  J ( \{
% \sum_{\begin{subarray}~q=\mathcal{N}(p,i)\\i\in~[1,N]\end{subarray}}
% (\mathbf{X}^{S,i\Delta\theta_\mathrm{o}}_q + \mathbf{X}^{1-S,i\Delta\theta_\mathrm{b}}_q)k_{pq} \},y )~~~~~~\\
% \mathrm{~~subject~to~~} \forall p, \|\mathbf{k}_p\|_{0} \leq\epsilon, \max(\mathbf{k}_p)=k_{pp},\sum_{q\in\mathcal{N}(p)}k_{pq}=1~\nonumber \\
% \forall p,q, \mathbf{k}_p=\mathbf{k}_q, ~\mathrm{if}~\mathbf{S}(p)=\mathbf{S}(q),
% \|\theta_\mathrm{o}\|_{\infty} \leq\epsilon_\theta, \|\theta_\mathrm{b}\|_{\infty}\leq\epsilon_\theta. \nonumber
% \end{align}
% }
%---------------------------
% 
\begin{wrapfigure}{l}{.55\textwidth} %.61
% {\footnotesize
{\scriptsize
\vspace{-1.8em}
\begin{align}\label{eq:motion_adv_obj}
\argmax_{\mathcal{K},\theta_\mathrm{o},\theta_\mathrm{b}}  J ( \{
\sum_{\begin{subarray}~q=\mathcal{N}(p,i)\\i\in[1,N]\end{subarray}}
(\mathbf{X}^{S,i\Delta\theta_\mathrm{o}}_q + \mathbf{X}^{1-S,i\Delta\theta_\mathrm{b}}_q)k_{pq} \},y )\\
\mathrm{~subject~to~} \forall p, \|\mathbf{k}_p\|_{0} \leq\epsilon, \max(\mathbf{k}_p)=k_{pp},\sum_{q\in\mathcal{N}(p)}k_{pq}=1\nonumber\\
\forall p,q, \mathbf{k}_p=\mathbf{k}_q, ~\mathrm{if}~\mathbf{S}(p)=\mathbf{S}(q),
\|\theta_\mathrm{o}\|_{\infty} \leq\epsilon_\theta, \|\theta_\mathrm{b}\|_{\infty}\leq\epsilon_\theta. \nonumber
\end{align}}
\vspace{-2.7em}
\end{wrapfigure}
%---------------------------
where $\epsilon_\theta\in [0,1]$ controls the maximum translations of the object/background. Here, we use the spatial transformer network \cite{JaderbergNIPS2015} to perform translation according to $\theta_{\mathrm{o}}$ and $\theta_{\mathrm{b}}$, enabling the gradient propagate to all kernels. There are two main differences about the constrained terms \cf Eq.~(\ref{eq:kernel_adv_obj}): (1) The translation parameters are added to guide the generation of the adversarial example; (2) The kernels are set to be the same within the same region, which is needed to generate visually natural motion blur, since pixels in the object region usually have the same motion. As shown in Fig.~\ref{fig:kpn}, by incorporating saliency and motion regularization, the AB$\kern-3pt$\textcolor{mygray}{B}A's adversarial example looks visually more natural than the one by AB$\kern-3pt$\textcolor{mygray}{B}A$_\text{pixel}$.

\iffalse
The attack method can be integrated into any gradient-based additive-perturbation attack methods, \eg, FGSM, BIM, MIFGSM, \etc. In this work, we adopt the MIFGSM for the high transferability across models which is a significant property for real-world security application.\if 0 \fei{The previous sentence is too high level. We should claimed that it is general that The kernel and translation can be calculated by different attack/optimization techniques (\eg, FGSM, MIFGSM)}\fi There are two main differences about the constrained terms \cf Eq.~(\ref{eq:motion_adv_obj}): 1) The translation parameters are added to guide the generation of the adversarial example. 2) The kernels are set to be the same within the same region, which is significant to generate visually natural motion blur, since pixels in the object region usually have the same motion.
\fi

\vspace{-0.7em}
%----------------------------------------------------------------------
\subsubsection{Attacking Algorithm}\vspace{-0.5em}
\label{subsubsec:algorithm}
%
% \qing{Please help check the following steps.}
We summarize the workflow of our attacking algorithm in the following steps: 1) Calculate the saliency map of an image, \ie, $\mathbf{X}^\text{real}$, via PoolNet and obtain $\mathbf{S}$. 2) Initialize $\theta_{\mathrm{o},t}=\theta_{\mathrm{b},t}=[0,0]$ and set each kernel $\mathbf{k}_{p,t}$ of $\mathcal{K}_t$ by $\{k_{pp,t}=1,k_{pq,t}=0|\forall q\in\mathcal{N}(p), q\neq p\}$ where $t=0$ denotes the first iteration. 3) Calculate $\mathbf{X}^\text{adv}_{t}$ via Eq.~(\ref{eq:motion_adv}), which is also visualized in Fig.~\ref{fig:method} for better understanding. 4) Calculate the gradient of $\mathbf{X}^\text{adv}_t$ with respect to the objective function and obtain $\nabla_{\mathbf{X}^\text{adv}_t}{J(\mathbf{X}^\text{adv}_t,y)}$. 5) Propagate the gradient through the spatial transformer network and obtain the gradients of $\mathcal{K}_t$, $\theta_{\mathrm{o},t}$, and $\theta_{\mathrm{b},t}$, \ie,   $\nabla_{\mathcal{K}_t}{J(\mathbf{X}^\text{adv}_t,y)}$, $\nabla_{\theta_{\mathrm{o},t}}{J(\mathbf{X}^\text{adv}_t,y)}$ and $\nabla_{\theta_{\mathrm{b},t}}{J(\mathbf{X}^\text{adv}_t,y)}$. 6) Update $\mathcal{K}_t$, $\theta_{\mathrm{o},t}$, and $\theta_{\mathrm{b},t}$ with a step size. 7) Update $t=t+1$ and go to the Step 3) for further optimization until it reaches the maximum iteration or $\mathbf{X}^\text{adv}_t$ fools the DNN. We will detail our settings in Sec.~\ref{subsec:settings}.
%When the maximum iteration is reached, we obtain $\mathbf{X}^\text{adv}$,  estimated kernels $\mathcal{K}^{*}$, and the translation parameters ($\theta_o^{*}$ and $\theta_b^{*}$).
%
%The attack method can be integrated into any gradient-based additive-perturbation attack methods, \eg, FGSM, BIM, MIFGSM. In this work, we not only show the effectiveness of the attack but also evaluate the transferability across models. Hence, we equip our method in MIFGSM which is more effective in achieving high transferability across models. 
%
%Then, we propagate the gradient through the spatial transformer network and obtain the gradient $\{\nabla_{\mathbf{k}_p}{J(\mathbf{X}^\text{adv}_t,y)}\}$, $\nabla_{\theta_\mathrm{o}}{J(\mathbf{X}^\text{adv}_t,y)}$ and $\nabla_{\theta_\mathrm{b}}{J(\mathbf{X}^\text{adv}_t,y)}$. The kernel and translation parameters are updated for realizing iterative attack.

\vspace{-0.7em}
\subsection{AB$\kern-3pt$\textcolor{mygray}{B}A$_\text{physical}$: Towards Real-World Adversarial Blur Attack }\vspace{-0.5em}
\label{subsec:pysical}
As introduced in Sec.~\ref{subsubsec:algorithm}, AB$\kern-3pt$\textcolor{mygray}{B}A takes a real example as the input and produces an adversarial blur example, the estimated kernels ($\mathcal{K}^{*}$), and translation parameters ($\theta_o^{*}$ and $\theta_b^{*}$). Then, it posts an interesting problem whether we could use the estimated translation parameters to guide camera or object moving, in generating a real-world adversarial blur examples. There are three main challenges: 1) We cannot control kernels' values in the real world. The optimized kernels are to let the adversarial examples fool deep models and may not exist in the real world. 2) It is difficult to precisely control the object or camera's moving without a high-precision robot arm. 3) To transfer the image translation to camera translation, we need know the object depth and camera intrinsic parameters. 

To alleviate these challenges, we conduct the following modifications of our AB$\kern-3pt$\textcolor{mygray}{B}A: 1) we fix the kernels $\mathcal{K}$ to be average kernels (\ie, we set each kernel's elements as $\frac{1}{N}$ where $N$ is the kernel size) that let the generated adversarial blur follow the real-world motion blur\footnote{\tiny We use the average kernels to simulate the real motion blur generation process, which has been used to construct training dataset for deblurring methods \cite{Nah2017CVPR,Noroozi2017}.}. 2) We force the object and background to share the translation parameters. As a result, AB$\kern-3pt$\textcolor{mygray}{B}A simulates the blur generation through camera moving, where object and background have the same motion thus we can produce blurred examples by moving the camera. 3) In the real world, we could use a RGB-D camera to get an object's depth and leverage a calibration software to obtain the intrinsic parameters of the camera.

Based on these, we can generate a real-world adversarial blur example by: 1) capturing a picture containing an object in a real-world scene. 2) using our AB$\kern-3pt$\textcolor{mygray}{B}A to calculate $\theta_o^{*}$ or $\theta_b^{*}$. 3) calculating the camera translations according to object depth and camera intrinsic parameters. 4) moving camera according to the camera translations and take a real blur picture during the moving process as the output. In particular, we will validate AB$\kern-3pt$\textcolor{mygray}{B}A$_\text{physical}$ in  Sec. \ref{sec:exp_real_world} through the AirSim simulator \cite{airsim2017fsr} within which we can control a simulated camera precisely and obtain the depth map. We also conduct an experiment with a mobile phone to preliminarily verify our method in the real world.

%----------------------------------------------------------------------
%----------------------------------------------------------------------
\vspace{-1em}
\section{Experimental Results}
% \vspace{-1em}

% In this section, we conduct comprehensive experiments to demonstrate the effectiveness of our proposed method, by investigating the following five research questions: 1) Is the transferability of our proposed method across models comparable or even better than SOTA attack methods?  2) Could our method produce visually natural examples? 3) Could we generate adversarial blurred examples that possibly occur in the real world? 4) As a blurring method, could SOTA deblurring methods easily defend our attack? 5) How is attacking success rate affected by hyper-parameters, \eg, $\epsilon$, $\epsilon_\theta$, motion direction, and blur region? 
% if state-of-the-art deblurring methods could defend it easily?

%----------------------------------------------------------------------
\vspace{-0.7em}
\subsection{Experimental Settings}\vspace{-0.5em}
\label{subsec:settings}
{\bf Dataset and Models.} We use NeurIPS'17 adversarial competition dataset \cite{KurakinNIPS2017cp}, compatible with ImageNet, for all the experiments.
To validate our method's transferability, we consider four widely used models, \ie, Inception v3 (Inc-v3) \cite{SzegedyCVPR2016}, Inception v4 (Inc-v4), Inception ResNet v2 (IncRes-v2) \cite{SzegedyAAAI2017}, and Xception \cite{Chol17Xception}. We further compare on four defense models: Inc-v3$_\text{ens3}$, Inc-v3$_\text{ens4}$, and IncRes-v2$_\text{ens}$ from \cite{TramerICLR2018} and high-level representation guided denoiser (HGD) \cite{LiaoCVPR2018} with the highest ranking in NeurIPS'17 defense competition.
We report the results of Inc-v3 in the paper, and put more results of other models (\eg, Inc-v4, IncRes-v2, Xception) in the supplementary material.
%
% All adversarial examples are generated from Inc-v3 treating it as a whitebox model. We reports the results of using other whitebox models, \eg, Inc-v4, IncRes-v2, and Xception, in the supplementary material.

\noindent{\bf Baselines.} We consider two kinds of baselines. The first scope is SOTA additive-perturbation-based attacks, \eg, FGSM \cite{GoodfellowARXIV2014}, MIFGSM \cite{DongCVPR2018}, DIM \cite{XieCVPR2019}, TIFGSM, TIMIFGSM, and TIDIM \cite{DongCVPR2019}\footnote{\tiny We use the released code in \url{github.com/dongyp13/Translation-Invariant-Attacks} to get the results of FGSM, MIFGSM, DIM, TIFGSM, TIMIFGSM, and TIDIM.}, and interpretation-based noise \cite{FongICCV2017}. 
%The basic iterative method \cite{KurakinICLR2017} and the C\&W method \cite{CW_2017_SSP} are not included, since they are not good at generating transferable adversarial examples \cite{DongCVPR2019,DongCVPR2018}. 
The second kind contains three blur-based methods including the interpretation-based blur \cite{CW_2017_SSP}, Gaussian blur \cite{rauber2017foolbox}, and Defocus blur. 
For the transferability comparison in Sec.~\ref{subsec:exp_comp_trans}, we follow the default settings in \cite{DongCVPR2019} for the first group attacks.
% \ie, the maximum perturbation is to be $16$ with pixel values in [0,255], the weight decay factor and the step size for MIFGSM and TIMIFGSM are set as 1.0 and 1.6, and the transformation probability of DIM and TIDIM is set to 0.7, to achieve high transferability. \qing{put it to supplementar}
For all iterative attack methods including ours, we set the iteration number to be 10.
For blur-based baselines, we set the standard variation of Gaussian blur and the kernel size of Defocus blur to be 15.0, which is the same with our method for a fair comparison. 
For the image quality comparison in Sec.~\ref{subsec:exp_comp_trans}, we tune the hyper-parameter of all attacks to cover the success rate from low to high and show the relationship between image quality and the success rate. This helps to see if our method could maintain high attack success rate and transferability while keeping the image natural.

\noindent{\bf Setup of AB$\kern-3pt$\textcolor{mygray}{B}A.} %
In the experimental part of Sec.~\ref{subsec:exp_comp_trans}, we implement our methods by setting the hyper-parameters, \ie, maximum translation $\epsilon_\theta$ to 0.4 and maximum valid kernel size $\epsilon$ to 15.0 with the iteration 10. The step sizes are set to $0.04$ and $1.5$ for updating the kernels and translation parameters, respectively. Such setups are around the medium values among the range of our hyper-parameters, well balancing the attack success rate and visual quality. We discuss the effect of $\epsilon_\theta$ and $\epsilon$ in Sec.~\ref{subsec:analysis}. In addition, we have four variants of our method, \ie, { AB$\kern-3pt$\textcolor{mygray}{B}A}$_\text{pixel}$, { AB$\kern-3pt$\textcolor{mygray}{B}A}$_\text{obj}$, { AB$\kern-3pt$\textcolor{mygray}{B}A}$_\text{bg}$, and { AB$\kern-3pt$\textcolor{mygray}{B}A}$_\text{image}$. The first one is the attack introduced in Sec.~\ref{subsec:analysis} while the rest three methods blur different regions of input images, \eg, AB$\kern-3pt$\textcolor{mygray}{B}A$_\text{obj}$ only adds motion blur to the object region by fixing the kernels of background pixels to be  $\{k_{pp}=1,k_{pq}=0|\mathbf{S}(p)=0,q\in\mathcal{N}(p), q\neq p\}$ and AB$\kern-3pt$\textcolor{mygray}{B}A$_\text{image}$ adds motion blur to the whole image while forcing object and background to share the kernels and translations.

\noindent{\bf Metrics.} We use the success rate (Succ. Rate), \ie, the rate of adversarial examples that fool attacked DNNs, to evaluate the effectiveness of the attacks.
Regarding quality of generated adversarial examples,
we use 
% To evaluate the imperceptibility of succeed adversarial examples, we use a general image quality assessment, \ie,
BRISQUE \cite{Mittal2012} instead of $L_\mathrm{p}$ norms,
% , $L_2$, and $L_\infty$ of perturbations
PSNR, or SSIM due to the following reasons: 1) our AB$\kern-3pt$\textcolor{mygray}{B}A cannot be fairly evaluated by the $L_1$, $L_2$, $L_\infty$, PSNR, or SSIM metrics since the perturbation is not additive and no longer well aligned pixel-to-pixel. 2) BRISQUE \cite{Mittal2012} is a natural scene statistic-based distortion-generic blind/no-reference (NR) image quality assessment and widely used to evaluate losses of `naturalness' in the image due to the presence of distortions including additive noise and blur. Let it be noted that a more natural image often has a smaller BRISQUE value.

%--------------------------
\begin{SCtable}
%\begin{wraptable}{r}{0.6\columnwidth}
\scalebox{0.7}{
	\caption{Adversarial comparison results on NeurIPS'17 adversarial competition dataset according to the success rate. The adversarial examples are generated from Inc-v3. There are two comparison groups. For the first one, we compare  blur-based methods, \ie, Interpretation-based blur (Interp$_\text{blur}$), GaussBlur, and DefocusBlur with our {\bf AB$\kern-3pt$\textcolor{mygray}{B}A} by considering the effects of attacking different regions, \ie, object or background regions, of inputs. In addition to above methods, the second group comparison contains additive-perturbation-based attacks, \ie, Interpretation-based noise (Interp$_\text{noise}$) \cite{FongICCV2017}, FGSM \cite{GoodfellowARXIV2014}, MIFGSM \cite{DongCVPR2018}, DIM \cite{XieCVPR2019}, and TIFGSM, TIMIFGSM, and TIDIM \cite{DongCVPR2019}. We highlight the top three results with \first{pink}, \second{yellow}, and \third{blue}, respectively.}
	\label{tab:attack_results}
	%\vspace{-0em}
	\tiny
% 	\begin{center}
	\setlength\tabcolsep{1.5pt} % default value: 6pt
		\begin{tabular}{l|c|c|c|c|c|c|c|c}
			\hline
% 			\multirow{}{}{} 
			\rowcolor{tabgray2}~ & \multicolumn{4}{c|}{Attacking Results (Inc-v3)} & \multicolumn{4}{c}{Defence Results (Inc-v3)} \\
			\cline{2-9}
			\rowcolor{tabgray2} & Inc-v3 & Inc-v4 & IncRes-v2 & Xception & Inc-v3$_\text{env3}$ & Inc-v3$_\text{env4}$ & IncRes-v2$_\text{ens}$ & HGD \\
			\hline
			\hline
			{GaussBlur} & 34.7 & 22.7 & 18.4 & 26.1 & 23.6 & 23.8 & 19.3 & 16.9 \\
		    \cline{2-9}
			{GaussBlur}$_\text{obj}$ & 13.6 & 6.0 & 5.2 & 7.1 & 8.6 & 7.8 & 6.3 & 4.6 \\
			\cline{2-9}
			{GaussBlur}$_\text{bg}$ & 18.8 & 10.8 & 9.2 & 12.0 & 13.0 & 13.1 & 10.9 & 8.7 \\
		    \hline
			{DefocBlur} & 30.0 & 16.8 & 11.1 & 18.8 & 17.5 & 18.3 & 15.0 & 12.9 \\
		    \cline{2-9}
			{DefocBlur}$_\text{obj}$ & 10.0 & 3.0 & 2.9 & 3.6 & 5.2 & 4.6 & 3.8 & 2.7 \\
			\cline{2-9}
			{DefocBlur}$_\text{bg}$ & 16.9 & 9.2 & 7.0 & 10.5 & 10.1 & 10.3 & 9.2 & 7.8 \\
			\hline
			{Interp}$_\text{blur}$ & 34.7 & 3.6 & 0.5 & 3.4 & 7.1 & 7.1 & 4.3 & 1.4 \\
			\hline
			\hline
			\rowcolor{tabgray1}{ AB$\kern-3pt$\textcolor{mygray}{B}A}$_\text{obj}$ & 21.0 & 4.9 & 4.2 & 7.0 & 10.1 & 10.5 & 8.3 & 4.9 \\
			\cline{2-9}
			\rowcolor{tabgray1}{ AB$\kern-3pt$\textcolor{mygray}{B}A}$_\text{bg}$ & 30.9 & 11.6 & 10.1 & 12.9 & 1.2 & 0.8 & 1.2 & 0.5 \\
			\cline{2-9}
			\rowcolor{tabgray1}{ AB$\kern-3pt$\textcolor{mygray}{B}A}$_\text{image}$ & 62.4 & 29.8 & 28.8 & 34.1 & 43.2 & 43.8 & \cellcolor{top3} 38.9 & 28.4 \\
			\cline{2-9}
			\rowcolor{tabgray1}{ AB$\kern-3pt$\textcolor{mygray}{B}A}$_\text{pixel}$ & 89.2 & \cellcolor{top3} 65.5 & \cellcolor{top3} 65.8 & \cellcolor{top2} 71.2 & \cellcolor{top1} 69.8 & \cellcolor{top1} 72.5 & \cellcolor{top1} 68.0 & \cellcolor{top1} 63.1 \\
			\cline{2-9}
			\rowcolor{tabgray1}{ AB$\kern-3pt$\textcolor{mygray}{B}A} & 65.6 & 31.2 & 29.7 & 33.5 & \cellcolor{top3} 46.6 & \cellcolor{top2} 48.7 & \cellcolor{top2} 41.2 & \cellcolor{top3} 31.0 \\
			\hline
			\hline
			{Interp}$_\text{noise}$ & 95.8 & 20.5 & 15.6 & 22.9 & 16.8 & 16.1 & 9.4 & 3.3 \\
		    \hline
			{FGSM} & 79.6 & 35.9 & 30.6 & 42.1 & 15.6 & 14.7 & 7.0 & 2.1 \\
			\hline
			{MIFGSM} & \cellcolor{top3} 97.8 & 47.1 & 46.4 & 47.7 & 20.5 & 17.4 & 9.5 & 6.9 \\
			\hline
		    {DIM} &  \cellcolor{top2} 98.3 & \cellcolor{top2} 73.8 & \cellcolor{top2} 67.8 & \cellcolor{top1} 71.6 & 24.2 & 24.3 & 13.0 & 9.7 \\
			\hline
			{TIFGSM} & 75.4 & 37.3 & 32.1 & 38.6 & 28.2 & 28.9 & 22.3 & 18.4 \\
			\hline
			{TIMIFGSM} & 97.9 & 52.4 & 47.9 & 44.6 & 35.8 & 35.1 & 25.8 & 25.7 \\
			\hline
			{TIDIM} & \cellcolor{top1} 98.5 &  \cellcolor{top1} 75.2 & \cellcolor{top1} 69.2 & \cellcolor{top3} 61.3 & \cellcolor{top2} 46.9 & \cellcolor{top3}  47.1 & 37.4 & \cellcolor{top2}  38.3  \\
			\hline
		\end{tabular}
		}
% 	\end{center}
	\vspace{-3em}
\end{SCtable}
%\end{wraptable}
%--------------------------

%----------------------------------------------------------------------
\vspace{-0.7em}
\subsection{Comparison with Baselines on Transferability}\vspace{-0.5em}
\label{subsec:exp_comp_trans}

Tab.~\ref{tab:attack_results} summarizes the comparison results and we discuss from two aspects: 1) the comparison with additive-perturbation-based attacks. 2) the advantages over blur-based methods. 
For the first aspect, compared with most of the additive-perturbation-based attacks, our method, \ie, AB$\kern-3pt$\textcolor{mygray}{B}A and {AB$\kern-3pt$\textcolor{mygray}{B}A}$_\text{pixel}$, achieve much higher success rate on all defences models, demonstrating the higher transferability of AB$\kern-3pt$\textcolor{mygray}{B}A over baselines. Compared with two SOTA methods, \ie, DIM and TIDIM, AB$\kern-3pt$\textcolor{mygray}{B}A$_\text{pixel}$ achieves slightly lower success rate when attacking Inc-v4 and IncRes-v2 while obtaining higher results than TIDIM when attacking Xception. In summary, our methods {AB$\kern-3pt$\textcolor{mygray}{B}A}$_\text{pixel}$ and {AB$\kern-3pt$\textcolor{mygray}{B}A} have competitive transferability with SOTA additive-perturbation-based attacks while achieving significant advantages in attacking defence models.
For the second aspect, AB$\kern-3pt$\textcolor{mygray}{B}A achieves higher success rate than GaussBlur, DefocusBlur, and Interp$_\text{blur}$ on all normally trained and defense models. We also compare them for attacking object or background regions. Obviously, the success rates of all methods decrease significantly when we add blur to only the object or background regions. 
% Moreover, the decreasing values of our method are larger than others, which means our method relies on more image information to realize effective attacking.
%
Besides these transferability results, we also add an analysis in the supplementary material about the interpretable explanation of the high transferability of our method by comparing with FGSM and MIFGSM. 

\vspace{-0.7em}
\subsection{Comparison with Baselines on Image Quality}\vspace{-0.5em}

We conduct an analysis about the attack success rate and image quality (\ie, measured by BRISQUE \cite{Mittal2012} that evaluates losses of ‘naturalness’ in the image). Note that, smaller BRISQUE corresponds to more natural images. We also consider two comparison groups: 1) the blur-based attacks, \ie, GaussBlur and DefocBlur, and 2) SOTA additive-perturbation-based attacks, \ie, FGSM, MIFGSM, DIM, TIFGSM, TIMIFGSM, and TIDIM. For each compared method, we tune their hyper-parameters
%
% \footnote{\tiny For FGSM, MIFGSM, DIM, TIFGSM, TIMIFGSM, and DIM, we tune their success rate by changing their maximum perturbation, \ie, $\epsilon_\mathrm{a}$ in Eq.~(\ref{eq:add_adv_obj}).
% % from 1 to 55.
% We tune the blur-based attacks by tuning the kernel size from 3 to 21. For AB$\kern-3pt$\textcolor{mygray}{B}A, we change the $\epsilon$ and $\epsilon_\theta$ from 5 to 50 and 0.1 to 1, respectively.} 
%
to cover success rates from low to high on the NeurIPS'17 adversarial competition dataset. For each hyper-parameter, we calculate the average BRISQUE of the adversarial images that successfully fool DNNs. 
As a result, for each attacked model, we can draw a plot for an attack method and the points more near the top left corner are better (\ie, looks more naturally while fooling DNNs).
%
%---------------------------
\begin{wrapfigure}{r}{0.48\columnwidth}
\vspace{-1em}
% \begin{figure}[t]
\centering
\includegraphics[width=0.48\columnwidth]{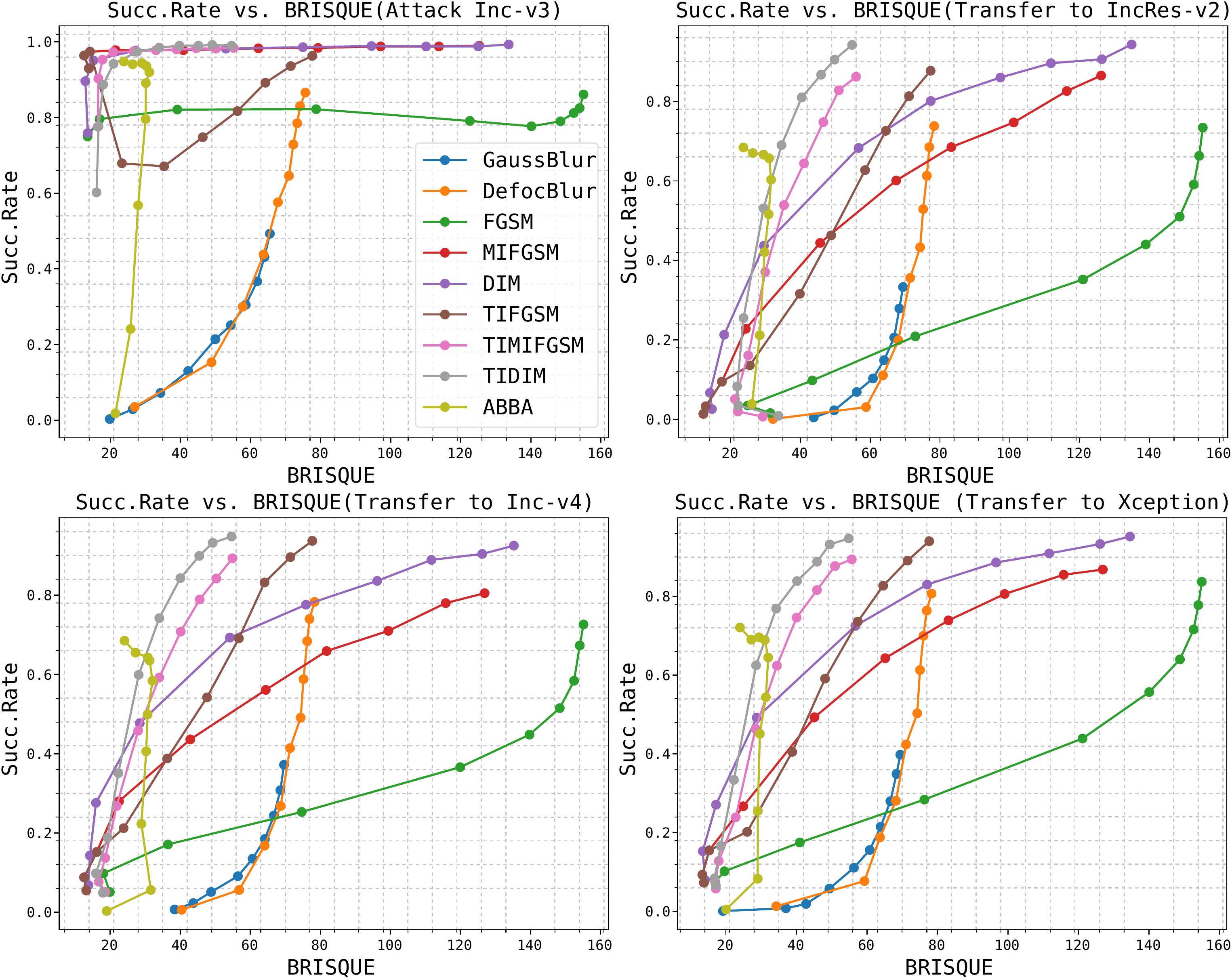}
\caption{ Succ. Rate vs. BRISQUE. Note that, smaller BRISQUE corresponds to more natural images. 
%For each compared method, we have 10 hyper-parameters to get 10 different success rates on the NeurIPS'17 adversarial competition dataset. For each hyper-parameter, we calculate the average BRISQUE of adversarial images that successfully fool deep models. For each attacked model, we can draw a plot for each attack method. The adversarial examples are generated from Inc-v3. The sub-figures from left to right are results of attacking Inc-v3, Inc-v2, IncRes-v4, and Xception, respectively.
%\felix{emphasize top-left is BEST zone}
}
\label{fig:comp_quality}
% \end{figure}
\vspace{-1em}
\end{wrapfigure}
%---------------------------
%
As shown in Fig.~\ref{fig:comp_quality}, in general, the quality of images generated by all baseline methods gradually gets worse as their success rate becomes larger. In contrast, the BRISQUE of our AB$\kern-3pt$\textcolor{mygray}{B}A always stay at a small value even if the success rate increases, demonstrating that AB$\kern-3pt$\textcolor{mygray}{B}A can produce visually natural adversarial examples with high attack success rate. 
% When considering the whitebox attack shown at the first sub-figure, we find that additive-perturbation-based attacks are easier to reach high attack success rate than blurred-based attacks. 
%
% However, 
When transferring the adversarial examples to other models, we observe that the success rate of additive-perturbation-based attacks, especially the FGSM, MIFGSM, TIFGSM, and DIM, decrease sharply, while the blur-based attacks are not impacted so largely. 
Comparing with SOTA methods (\ie, TIDIM and TIMIFGSM) on the transferability, we see that the adversarial examples of AB$\kern-3pt$\textcolor{mygray}{B}A and the two methods have similar BRISQUE scores when their success rate is small. When success rate of TIDIM and TIFGSM further increases, their BRISQUE values become smaller than ours.

\vspace{-1.1em}
\subsection{Adversarial Blur Examples in the Simulation World and Real World}\vspace{-0.5em}
\label{sec:exp_real_world}
{\bf Results in AirSim environment:} We validate our AB$\kern-3pt$\textcolor{mygray}{B}A$_\text{physical}$ on the AirSim simulator \cite{airsim_url1} that supports hardware-in-loop with cameras for physically and visually realistic simulations.
%
%-------------------
\begin{wrapfigure}{r}{0.5\columnwidth}
\vspace{-1.2em}
\includegraphics[width=0.5\columnwidth]{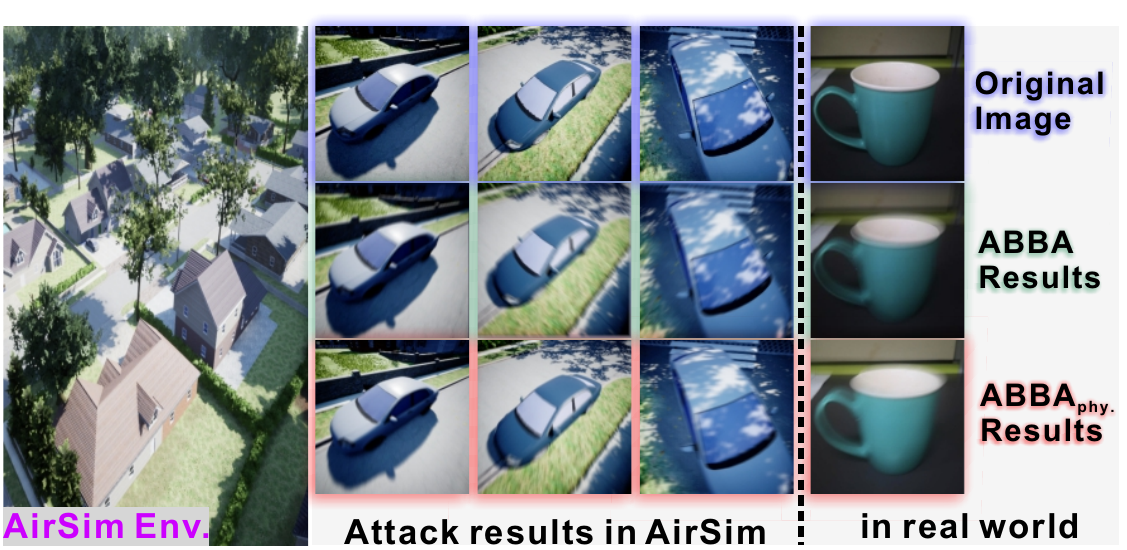}
\caption{Attack results in the AirSim environment (medium) and real world (right). The left sub-figure shows a snapshot of the neighborhood environment. The results of AB$\kern-3pt$\textcolor{mygray}{B}A$_\text{physical}$ are produced by physically moving the camera and mobile phone in the AirSim and real world, respectively. All the adversarial examples fool the Inc-v3 model.
}\vspace{-1.1em}
\label{fig:realvsadv}
\end{wrapfigure}
% \begin{SCfigure}
% \vspace{-0em}
% \includegraphics[width=0.55\columnwidth]{./fig_sim_real.pdf}
% \caption{Attack results in the AirSim environment (medium) and real world (right). The left sub-figure shows a snapshot of the neighborhood environment. The results of AB$\kern-3pt$\textcolor{mygray}{B}A$_\text{physical}$ are produced by physically moving the camera and mobile phone in the AirSim and real world, respectively. All the adversarial examples fool the Inc-v3 model.
% }\vspace{-2em}
% \label{fig:realvsadv}
% \end{SCfigure}
%-------------------
%
AirSim also provides APIs to retrieve relevant data (\eg, real-time depth map) and controls cameras in a platform independent way, which meets the requirements of  AB$\kern-3pt$\textcolor{mygray}{B}A$_\text{physical}$ introduced in Sec.~\ref{subsec:pysical}. In particular, we choose the open-source Neighborhood environment \cite{airsim_url2}, containing 70 cars with various styles. For each car, we select a observation view where a DNN (\eg, Inc-v3) could classify it correctly. Then, we conduct the AB$\kern-3pt$\textcolor{mygray}{B}A$_\text{physical}$ experiment for each car with the following steps: 1) Set a camera to capture an original image of the car at the selected observation view. 2) Use our method introduced in Sec.~\ref{subsec:pysical} to calculate an adversarial blurred image (\ie, AB$\kern-3pt$\textcolor{mygray}{B}A's result) and the camera translation with the depth map and camera intrinsic parameters.
3) Move the camera along a straight line to the translation destination while taking $N$ pictures and averaging them to the final blurred image~(\ie, AB$\kern-3pt$\textcolor{mygray}{B}A$_\text{physical}$'s result), which is equivalent to the motion blur generation process \cite{Nah2017CVPR}. 4) 
Test whether the real blurred image and adversarial blurred image could fool DNNs (\ie, Inc-v3, Inc-v4, IncRes-v2, and Xception in Table~\ref{tab:sim_results}) successfully.
We summarize the attack success rates in Table~\ref{tab:sim_results} and observe that: 1) AB$\kern-3pt$\textcolor{mygray}{B}A$_\text{physical}$ generated adversarial blur examples by physically moving cameras can achieve high success rate on Inc-v3, which demonstrates that there might exist real-world motion blur that can fool DNNs easily.
2) Although our AB$\kern-3pt$\textcolor{mygray}{B}A has very high transferability across DNNs, the physical adversarial examples cannot fool other DNNs easily. 
%
% \qing{Please check the following descriptions.} 
% \begin{SCtable}
% % \begin{wraptable}{r}{0.6\columnwidth}
% 	\caption{}%\vspace{-0em}
% 	\tiny
% % 	\begin{center}
% 	\setlength\tabcolsep{1.5pt} % default value: 6pt
% 		\begin{tabular}{l|c|c|c|c}
% 			\hline
% 			\rowcolor{tabgray2}~ & \multicolumn{4}{|c}{Adversarial examples from Inc-v3} \\
% 			\cline{2-5}
% 			\rowcolor{tabgray2} & Inc-v3 & Inc-v4 & IncRes-v2 & Xception \\
% 			\hline
% 			\hline
% 			{No. of Tested Cars} & 82 & 82 & 82 & 82\\
% 			{No. of Succ. AB$\kern-3pt$\textcolor{mygray}{B}A examples} & 80 & 72 & 68 & 74\\
% 			{No. of Succ. AB$\kern-3pt$\textcolor{mygray}{B}A$_\text{physical}$ examples} & 70 & 8 & 12 & 9\\
% 			\hline
% 		\end{tabular}
% % 	\end{center}
% 	\label{tab:attack_results}
%  	\vspace{-1em}
% \end{SCtable}
%
{\bf Results in the real world environment:} We further perform a preliminary experiment to validate our AB$\kern-3pt$\textcolor{mygray}{B}A$_\text{physical}$ in the real world through a mobile phone: 1) we capture a sharp image with a mobile phone, \ie, the cup in the first row of Fig.~\ref{fig:realvsadv}.
%----------------------------------------
\begin{wraptable}{r}{0.4\columnwidth}\vspace{-0em}
	\caption{Success rate of AB$\kern-3pt$\textcolor{mygray}{B}A and AB$\kern-3pt$\textcolor{mygray}{B}A$_\text{physical}$ for attacking DNNs in the AirSim environment. }
	\tiny
% 	\begin{center}
	\setlength\tabcolsep{1.5pt} % default value: 6pt
		\begin{tabular}{l|c|c|c|c}
			\hline
			\rowcolor{tabgray2}~ & \multicolumn{4}{|c}{Adversarial examples from Inc-v3} \\
			\cline{2-5}
			\rowcolor{tabgray2} & Inc-v3 & Inc-v4 & IncRes-v2 & Xception \\
			\hline
			\hline
			{Succ. Rate of AB$\kern-3pt$\textcolor{mygray}{B}A} & 97.6 & 87.8 & 82.9 & 90.2\\
			{Succ. Rate of  AB$\kern-3pt$\textcolor{mygray}{B}A$_\text{physical}$} & 85.3 & 9.7 & 14.6 & 11.0\\
			\hline
		\end{tabular}
% 	\end{center}
	\label{tab:sim_results}
 	\vspace{-1em}
\end{wraptable}
%----------------------------------------
%
2) we use AB$\kern-3pt$\textcolor{mygray}{B}A to generate an adversarial blur image with the Inc-v3 model, \ie, the second row of Fig.~\ref{fig:realvsadv}, and obtain the image translation parameters that indicate the mobile phone's moving direction and distance. 3) we move the mobile phone along the direction indicated by the translation parameters and shoot a real-blurred image in the same scene with longer exposure time. We find both blurred images are misclassified by the Inc-v3 model. Here, we ignore the cup's depth since we shoot the sharp and real-blurred images at almost the same position and empirically tune the moving distance of the mobile phone. Such operation could be replaced by high-precision robot arms in the future.

%----------------------------------------------------------------------
\vspace{-1em}
\subsection{Effect of Deblurring Methods}
\begin{wrapfigure}{r}{0.48\columnwidth} % Felix
% \begin{figure}[t]
\vspace{-1em}
\centering
\includegraphics[width=0.48\columnwidth]{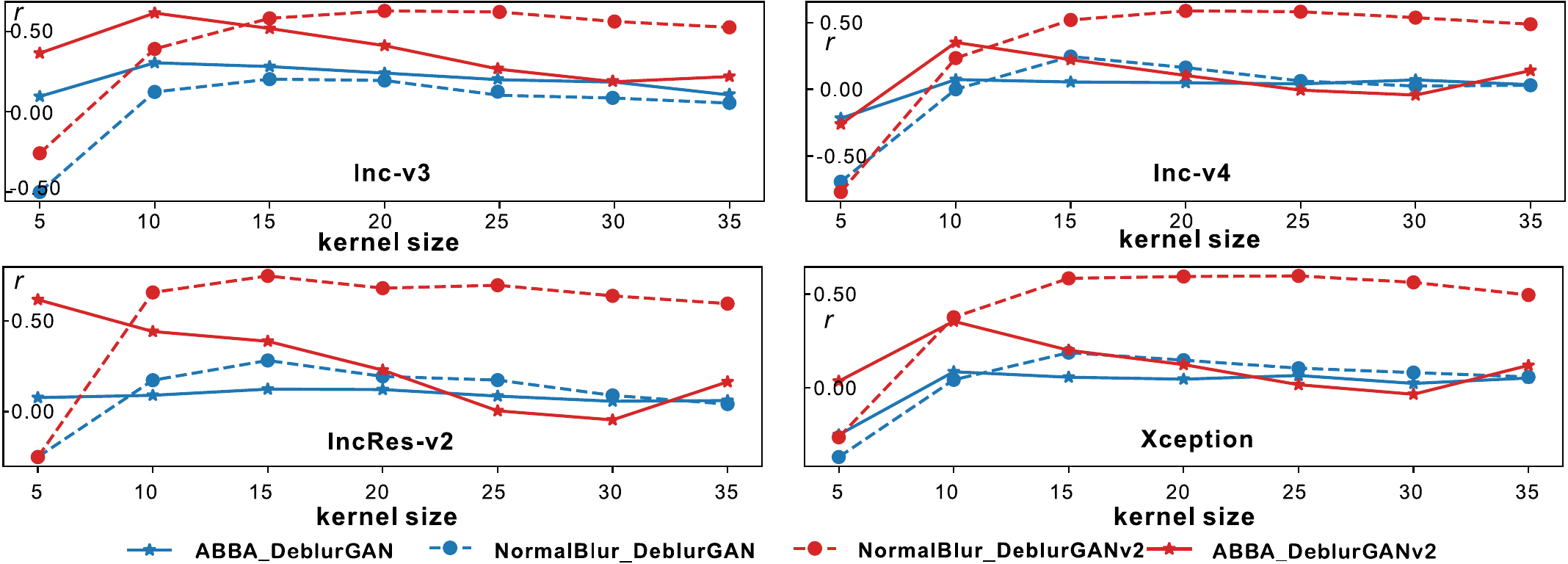}
\caption{The relative decrease of attack success rate before and after deblurring. Two state-of-the-art deblurring methods, \ie, DeblurGAN \cite{KupynCVPR2018} and DeblurGANv2 \cite{KupynICCV2019}, are used to deblur the our adversarial blurred examples and normal motion blurred images.}
\label{fig:deblur}
% \end{figure}
\vspace{-1em}
\end{wrapfigure}

\vspace{-0.5em}
Here, we discuss the effect of SOTA deblurring methods to our adversarial examples and `normal' motion blurred images. The `normal' motion blur is usually synthesized by averaging neighbouring video frames \cite{Nah2017CVPR}, which is equivalent to setting the kernel weights as $\frac{1}{N}$ to average the translated examples in Eq.~(\ref{eq:motion_adv}). We can regard such normal motion blur as an attack and add them to all images in the testing data. 
% The corresponding success rate can be calculated. %
We use DeblurGAN \cite{KupynCVPR2018} and DeblurGANv2 \cite{KupynICCV2019} to handle our adversarial examples and the normal motion blurred images and calculate the relative decrease of the success rate, \ie, $r=\frac{s-s'}{s}$ where $s$ and $s'$ represent the success rate before and after deblurring. 
Smaller $r$ means that the attack is more resilient against deblurring methods.
As shown in Fig.~\ref{fig:deblur}, in general, compared with the normal motion blurred images, our adversarial examples are harder to be handled by the state-of-the-art deblurring methods. 
For DeblurGANv2, the relative decrease $r$ of normal motion blur is usually larger than 0.5 when kernel size is in [15, 35]. This means that DeblurGANv2 can effectively remove the normal motion blur and improve the classification accuracy. In contrast, the $r$ of our method is always smaller than the $r$ of normal motion blur, and gradually decreases in the range of [15, 35], indicating that it is more difficult to use DeblurGANv2 to defend our attack as kernel size becomes larger.
Similar results are obtained on DeblurGAN where the difference of $r$ between our method and normal motion blur is much smaller than that on DeblurGANv2.

% When motion blur is slight~(\ie, kernel size equals to 5 or 10), we have $r<0$ for the normal motion blur, which means deblurring can improve its attack success rate.  

% The deblurring methods further decrease the success rate of slight normal motion blur~(\ie, the kernel size equals to 5 or 10) 

%----------------------------------------------------------------------
\vspace{-1.1em}
\subsection{Hyper-parameter Analysis and Ablation Study}\vspace{-0.5em}
\label{subsec:analysis}

{\bf Effect of $\epsilon$ and $\epsilon_\theta$:} We calculate the success rate of our method with different $\epsilon$ and $\epsilon_\theta$ in Eq.~(\ref{eq:motion_adv_obj}) and find that the success rates become gradually higher with the increase of $\epsilon$ and $\epsilon_\theta$. More results and discussions can be found in the supplementary material.
{\bf Effect of motion directions:} 
we fix $\epsilon_\theta$ and $\epsilon$ and tune the motion direction of object and background by setting different x-axis and y-axis translations to see the success rate variation. We find that the success rate reaches the highest value around $45^\circ$ moving direction. We visualize and discuss the results in the supplementary material.
%
%---------------------------
%---------------------------
%
%
{\bf Effect of blurred regions and importance of adaptive translations:} We conduct an ablation study by adding motion blur to different regions of an image. The success rate results are reported in Tab.~\ref{tab:attack_results}. Compared with AB$\kern-3pt$\textcolor{mygray}{B}A$_\text{pixel}$, 
AB$\kern-3pt$\textcolor{mygray}{B}A strikes a good balance between the attack success rate and visual effects.  Compared with other variants, AB$\kern-3pt$\textcolor{mygray}{B}A that jointly tunes the object and background translations can obtain much better transferability across normally trained and defense-based models. We show and discuss their visualization results in the supplementary material.

\section{Conclusions}\vspace{-1em}
%Motion blur is a common phenomenon in real-world image perception. 
\if 0
In this paper, we initiated the first step to comprehensively investigate the potential hazards of motion blur for DNNs. We proposed a new type of adversarial attack mode based on motion blur. %, named AB$\kern-3pt$\textcolor{mygray}{B}A.
We first propose the kernel-prediction-based attack that processes each pixel with a kernel that can be optimized by the supervision of the misclassification objective. Based on this, we further propose the saliency-regularized adversarial kernel prediction to make the motion-blurred image visually more natural and plausible. Comprehensive evaluation demonstrates the usefulness of our methods. Our results call for the attention of future research that takes motion blur effects into consideration during real-time image perception DNN designs.
We also hope our work facilities more general solutions of robust DNN towards addressing common camera motion blur effects.
\fi

In this paper, we have initiated the first step to comprehensively investigate the potential hazards of motion blur for DNNs.
We propose the kernel-prediction-based attack that can fool DNNs with a high success rate, which is further regularized by visual saliency to make the motion-blurred image visually more natural. Besides, we also validate that our adversarial blur might indeed exist in the real world. 
Comprehensive evaluation demonstrates the usefulness of our methods. Our results call for the attention of future research that takes motion blur effects into consideration during real-time image perception DNN designs.
We also hope that our work facilities more general solutions of robust DNN towards addressing common camera motion blur effects. 
Moreover, we have demonstrated that the proposed kernel-prediction-based adversarial attack can be extended to other kinds of attacking methods, \eg, adversarial attacks based on denoising \cite{cheng2020pasadena}, raining \cite{arxiv20_advrain}, camera exposure \cite{arxiv20_retinopathy}, and bias field in the medical imaging \cite{arxiv20_xray}. 
We will also discuss the effects of motion blur to visual object tracking \cite{guo2017learning,guo2017structure,guo2020selective}, through the proposed AB$\kern-3pt$\textcolor{mygray}{B}A and recent attacking method \cite{guo2020spark} against tracking.

%----------------------------------------------------------------------
%----------------------------------------------------------------------
\section{Broader Impact}\vspace{-1em}

In this work, we make an early attempt to investigate the motion-blur effects to DNNs, which is a common phenomenon in the real-world image capturing process of a camera.
We present the very \emph{first} attack based on manipulating the motion blur of the images. Through comprehensive experiments, we have demonstrated that very successful attacks can be well disguised in naturally-looking motion blur patterns of the image, unveiling the system vulnerabilities of any image capture stage that deals with camera or object motions.

Considering that image capturing and sensing is an integral and essential part of almost every computer vision application that interacts with the real world, the message we are trying to convey here is an important one, \ie, attackers can intentionally make use of motion blur, either by tampering with the camera sensing hardware or the image processing software to embed such an attack. Even unintentionally, the motion blur effects still commonly exist in the real-world application, posting threats to the DNNs behind the camera. 
This work is the first attempt to identify and showcase that such an attack based off image motion blur is not only feasible, but also leads to high attack success rate while simultaneously maintaining high realisticity in the image motion blur patterns. In a larger sense, this work can and will provide new thinking into how to better design the image capturing pipeline in order to mitigate potential risk caused by the vulnerabilities discussed herein, especially for mission- and safety-critical applications that are involved with moving objects or moving sensors such as autonomous driving scenarios, mobile face authentication with a hand-held device, computer-aided diagnostics in medical imaging, robotics, \etc.

Bad actors can potentially make use of this newly proposed attack mode as a wheel to pose risks on existing imaging systems that are not yet prepared for this new type of attack and effect based on image motion blur. We, as researchers, believe that our proposed method can accelerate the research and development of the DNN resilient mechanism against such motion blur effects. Therefore, our work can serve as an asset and a stepping stone for future-generation trustworthy design of computer vision DNNs and systems. 

In addition to the societal impact discussed above, the proposed method can also influence various research directions. For example, our proposed AB$\kern-3pt$\textcolor{mygray}{B}A method:

\begin{itemize}[leftmargin=*]
\item hints new data augmentation technique for training powerful DNN-based deblurring methods. 

\item hints new DNN design, detection/defense techniques to be resilience against motion blur effects. 

\item hints new direction of analyzing the effect of motion blur to video analysis tasks, \eg, real-time visual object detection, tracking, segmentation, and action recognition.

\end{itemize}

\begin{ack}\vspace{-1em}
% Use unnumbered first level headings for the acknowledgments. All acknowledgments
% More information about this disclosure can be found at: \url{https://neurips.cc/Conferences/2020/PaperInformation/FundingDisclosure}.
% Do {\bf not} include this section in the anonymized submission, only in the final paper. You can use the \texttt{ack} environment provided in the style file to autmoatically hide this section in the anonymized submission.
We appreciate the anonymous reviewers for their valuable feedback. This research was supported in part by Singapore National Cy-bersecurity R\&D Program No. NRF2018NCR-NCR005-0001, Na-tional Satellite of Excellence in Trustworthy Software System No.NRF2018NCR-NSOE003-0001, NRF Investigatorship No. NRFI06-2020-0022. It was also supported by JSPS KAKENHI Grant No.20H04168, 19K24348, 19H04086, and JST-Mirai Program Grant No.JPMJMI18BB, Japan, and the National Natural Science Foundation of China (NSFC) under Grant 61671325 and Grant 62072334. We gratefully acknowledge the support of NVIDIA AI Tech Center (NVAITC) to our research.
\end{ack}

%--------------------------------------------------------------
%--------------------------------------------------------------

\section{Supplementary Material}

In the main paper \footnote{\url{https://github.com/tsingqguo/ABBA}}, we have reported the attack results of Inc-v3 on four normal trained models and four defense models, and compared with 14 attack instances on the transferability and the image quality. In this supplementary material, (1) we present the evaluation details of our method regarding the transferability on five more defense models, comparing the visualization results with SOTA attacks, and discussing the attack results of another three DNNs. (2) We also conducted an in-depth hyper-parameter analysis and ablation study 
of our method, and posted an interpretable explanation about the difference between our method and baselines on the transferability. (3) We validated the generalization of our method by attacking an STN-based CNN. (4) We demonstrated that our method could help enhance the blur robustness of DNNs for the classification task with the results on ImageNetC. (5) We show more adversarial attack results in the real world. (6) We discuss the defense results via re-trained DeblurGANv2 with the blurred images from our methods.

Overall, the results of this supplementary material further demonstrated that the proposed adversarial blur attack can fool DNNs effectively while generating visually natural blurred images. All experimental results and discussions infer that motion blur as a common effect in the real world has a high risk of fooling SOTA DNNs and our attack methods initiate the first step to study the potential hazards of motion blur for DNNs.

%-----------------------------------------------------------------------
\subsection{Attack Results on Eight Defense Models}

Besides the results on the four defense models reported in the main paper, we also compared our method with baselines on another five defense models including R\&P~\cite{xie2018mitigating}, NeurIPS-r3\footnote{\url{https://github.com/anlthms/nips-2017}}, and three models from the stae-of-the-art feature denoise-based~(FD) defense method~\cite{Xie_2019_CVPR}~(\ie, ResNetXt101 with all denoising~(FD$_\text{R101}$), ResNet152 with four denoising blocks~(FD$_\text{R152}$), and adversarial trained baseline model ResNet152~(FD$_\text{R152B}$)). The R\&P method transforms input images through random resizing and padding, which ranked the second in the NeurIPS 2017 defense competition. NeurIPS-r3 is the third rank submission of NeurIPS 2017 defense competition and combines adversarial trained VGG16, Inc-v3, IncRes-v2, and ResNet152v2 models in an ensemble way. Besides, NeurIPS-r3 also performs transformations, \ie, shear, shifting, zoom, rotation, JPEG compression, and noise corruption, on input images. The FD method ranked the first in Competition on Adversarial Attacks and Defenses (CAAD)-2018.
%
% In the Table~1 of our main paper, we report the defense results of baselines, \ie, FGSM, MIFGSM, DIM, TIFGSM, TIMIIFGSM, and TIDIM that use $L_{2}$ norm bound and show low transferability. Here, we report the results of above baselines with $L_\infty$ norm bound and show the comparison results in Table~\ref{tab:defense_results}. 
%
Note that, all other results of above baselines in our main paper and supplementary material are based on $L_\infty$ norm bound.

As reported in Table~\ref{tab:defense_results}, our method, \ie,  AB$\kern-3pt$\textcolor{mygray}{B}A$_\text{pixel}$, achieves the highest transferability across all defense models and AB$\kern-3pt$\textcolor{mygray}{B}A has competitive results with the state-of-the-art baseline TIDIM. Such results reveal a potential big shortcoming of existing studies of defense methods, \ie, only considering the adversarial noise while ignoring other potential factors in physical environment. Note that, compared with adversarial noise, motion blur frequently happens in our daily life and widely exists among various computer vision-based applications, thus its influence to DNNs should be carefully studied and addressed.

%---------------------------
\begin{figure}[t]
\renewcommand\thefigure{I}
\centering
\includegraphics[width=0.65\columnwidth]{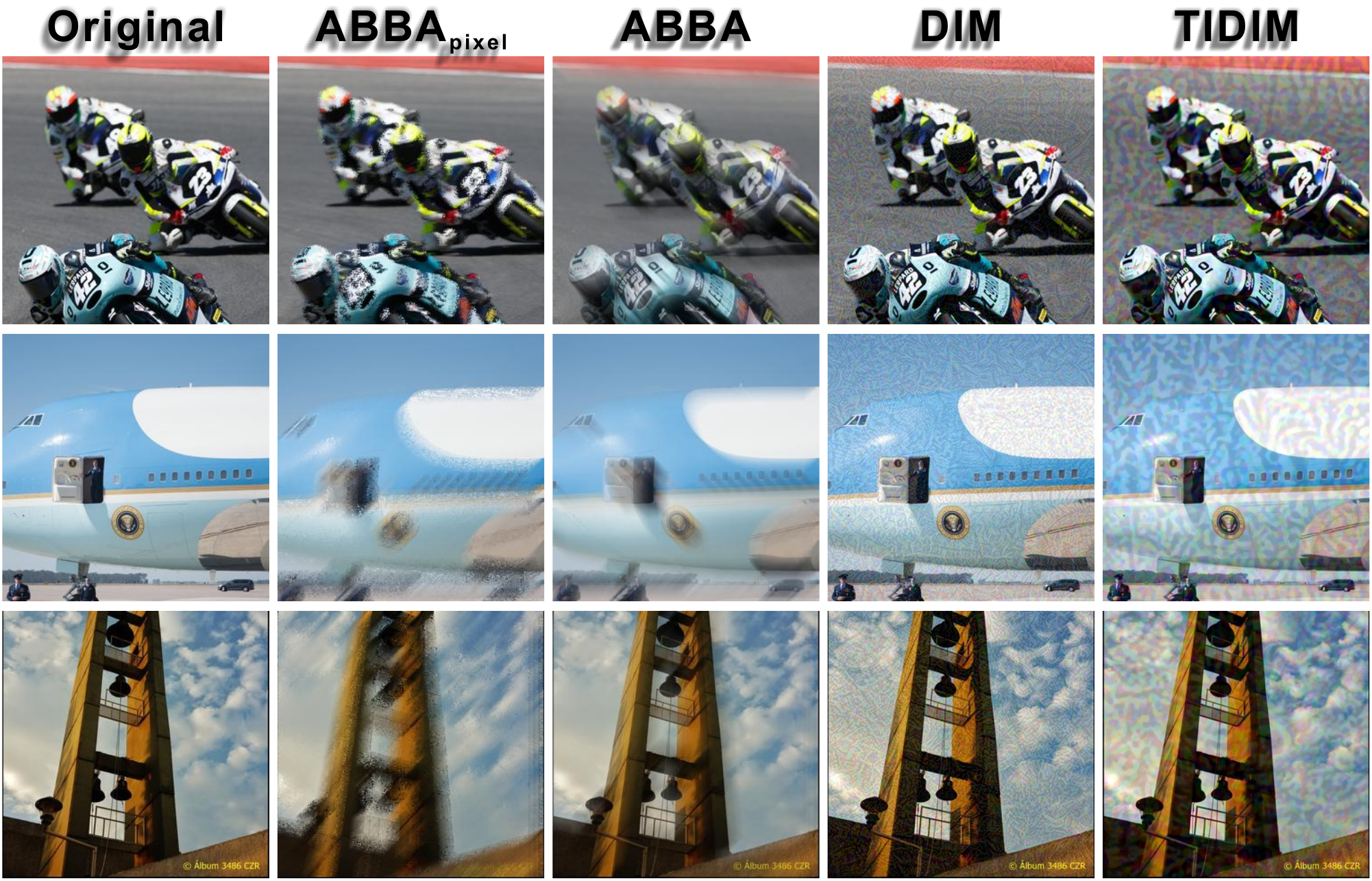}
\caption{Three visualization results of AB$\kern-3pt$\textcolor{mygray}{B}A$_\text{pixel}$, AB$\kern-3pt$\textcolor{mygray}{B}A, DIM, and TIDIM. All adversarial examples mislead the Inc-v3 model.}
\label{fig:cmp_vis0}
\vspace{-1.0em}
\end{figure}
%---------------------------

%--------------------------
\begin{table}[h]
\centering
\tiny
% \footnotesize
\setlength\tabcolsep{6pt} % default value: 6pt
% \scalebox{1.0}{
	\caption{Adversarial comparison results on NeurIPS'17 adversarial competition dataset according to the success rate. We use nine defense models to evaluate all attacks. The adversarial examples are generated from Inc-v3. There are two comparison groups. For the first one, we compare  blur-based methods, \ie, Interpretation-based blur (Interp$_\text{blur}$), GaussBlur, and DefocusBlur with our {\bf AB$\kern-3pt$\textcolor{mygray}{B}A} by considering the effects of attacking different regions, \ie, object or background regions, of inputs. In addition to above methods, the second group comparison contains additive-perturbation-based attacks, \ie, Interpretation-based noise (Interp$_\text{noise}$) \cite{FongICCV2017}, FGSM \cite{GoodfellowARXIV2014}, MIFGSM \cite{DongCVPR2018}, DIM \cite{XieCVPR2019}, and TIFGSM, TIMIFGSM, and TIDIM \cite{DongCVPR2019}. We highlight the top three results with \first{pink}, \second{yellow}, and \third{blue}, respectively.}%\vspace{-0em}
		\begin{tabular}{l|c|c|c|c|c|c|c|c|c}
			\hline
% 			\multirow{}{}{} 
			\rowcolor{tabgray2}~ & \multicolumn{9}{c}{Defence Results (Adv. Examples from Inc-v3)} \\
			\cline{2-10}
			\rowcolor{tabgray2} &~~Inc-v3$_\text{env3}$~~&~~Inc-v3$_\text{env4}$~~&~IncRes-v2$_\text{ens}$~&~~~~HGD~~~~&~~~~R\&P~~~~&NeurPIS-r3&~FD$_\text{R101}$~&~FD$_\text{R152}$~&~FD$_\text{R152B}$~\\
			\hline
			\hline
			{GaussBlur} & 23.6 & 23.8 & 19.3 & 16.9 & 17.2 & 17.6 & 35.6 & 35.8 & 35.9 \\
		    \cline{2-10}
			{GaussBlur}$_\text{obj}$ & 8.6 & 7.8 & 6.3 & 4.6 & 4.8& 5.1 & 13.9 & 13.9 & 14.6 \\
			\cline{2-10}
			{GaussBlur}$_\text{bg}$ & 13.0 & 13.1 & 10.9 & 8.7 & 10.0& 9.3 & 19.5 & 19.2 & 20.1 \\
		    \hline
			{DefocBlur} & 17.5 & 18.3 & 15.0 & 12.9 & 14.6 & 14.2 & 31.1 & 30.9 & 31.1 \\
		    \cline{2-10}
			{DefocBlur}$_\text{obj}$ & 5.2 & 4.6 & 3.8 & 2.7 & 3.3 & 2.9 & 10.8 & 10.5 & 11.1 \\
			\cline{2-10}
			{DefocBlur}$_\text{bg}$ & 10.1 & 10.3 & 9.2 & 7.8 & 9.0 & 8.1 & 19.5 & 17.6 & 18.5 \\
			\hline
			{Interp}$_\text{blur}$ & 7.1 & 7.1 & 4.3 & 1.4 & 2.9 & 2.9 & 25.5 & 25.8 & 28.6 \\
			\hline
			\hline
			\rowcolor{tabgray1}{ AB$\kern-3pt$\textcolor{mygray}{B}A}$_\text{obj}$ & 10.1 & 10.5 & 8.3 & 4.9 & 6.2& 7.1 & 18.7 & 18.4 & 19.1\\
			\cline{2-10}
			\rowcolor{tabgray1}{ AB$\kern-3pt$\textcolor{mygray}{B}A}$_\text{bg}$ & 1.2 & 0.8 & 1.2 & 0.5 & 0.6 & 0.7 & 43.5 & 44.1 & 45.5 \\
			\cline{2-10}
			\rowcolor{tabgray1}{ AB$\kern-3pt$\textcolor{mygray}{B}A}$_\text{image}$ &  43.2 &  43.8 & \cellcolor{top3} 38.9 & 28.4 & 34.1 & 35.0 & 61.1 & 61.9 & 62.4\\
			\cline{2-10}
			\rowcolor{tabgray1}{ AB$\kern-3pt$\textcolor{mygray}{B}A}$_\text{pixel}$ & \cellcolor{top1} 69.8 & \cellcolor{top1} 72.5 & \cellcolor{top1} 68.0 & \cellcolor{top1} 63.1 & \cellcolor{top1} 65.0 & \cellcolor{top1} 65.7 & \cellcolor{top1} 79.6 & \cellcolor{top1} 81.0 & \cellcolor{top1} 82.1 \\
			\cline{2-10}
			\rowcolor{tabgray1}{ AB$\kern-3pt$\textcolor{mygray}{B}A} & \cellcolor{top3} 46.6 & \cellcolor{top2} 48.7 & \cellcolor{top2} 41.2 & \cellcolor{top3} 31.0 & \cellcolor{top3} 36.7 & \cellcolor{top3} 38.5 & \cellcolor{top2} 64.2 & \cellcolor{top2} 64.6 & \cellcolor{top2} 65.6 \\
			\hline
			\hline
			{Interp}$_\text{noise}$ & 16.8 & 16.1 & 9.4 & 3.3 & 4.1 & 4.4 & 39.6 & 41.4 & 46.8\\
		    \hline
			{FGSM} & 15.6 & 14.7 & 7.0 & 2.1 & 6.5 & 9.8 & 39.2 & 41.4 & 45.3 \\
			\hline
			{MIFGSM} & 20.5 & 17.4 & 9.5 & 6.9 & 8.7 & 12.9 & 39.0 & 40.2 & 44.6 \\
			\hline
		    {DIM} & 24.2 & 24.3 & 13.0 & 9.7 & 13.3 & 18.0 & 39.1 & 40.3 & 45.1\\
			\hline
			{TIFGSM} & 28.2 & 28.9 & 22.3 & 18.4 & 19.8 & 24.5 & 39.7 & 41.8 & 45.4 \\
			\hline
			{TIMIFGSM} & 35.8 & 35.1 & 25.8 & 25.7 & 23.9 & 26.7 & 39.3 & 41.2 & 45.8\\
			\hline
			{TIDIM} &  \cellcolor{top2} 46.9 & \cellcolor{top3} 47.1 & 37.4 & \cellcolor{top2} 38.3 &  \cellcolor{top2} 36.8 &  \cellcolor{top2} 41.4 & \cellcolor{top3} 40.0 & \cellcolor{top3} 42.2 & \cellcolor{top3} 45.8\\
			\hline
		\end{tabular}
% 		}
	\label{tab:defense_results}
 	\vspace{-2em}
\end{table}
%--------------------------

%-----------------------------------------------------------------------
%-----------------------------------------------------------------------
\subsection{Visualization Comparison with Baselines}
\begin{figure}[h]
\vspace{-1em}
\renewcommand\thefigure{IX}
\centering
\includegraphics[width=1.0\columnwidth]{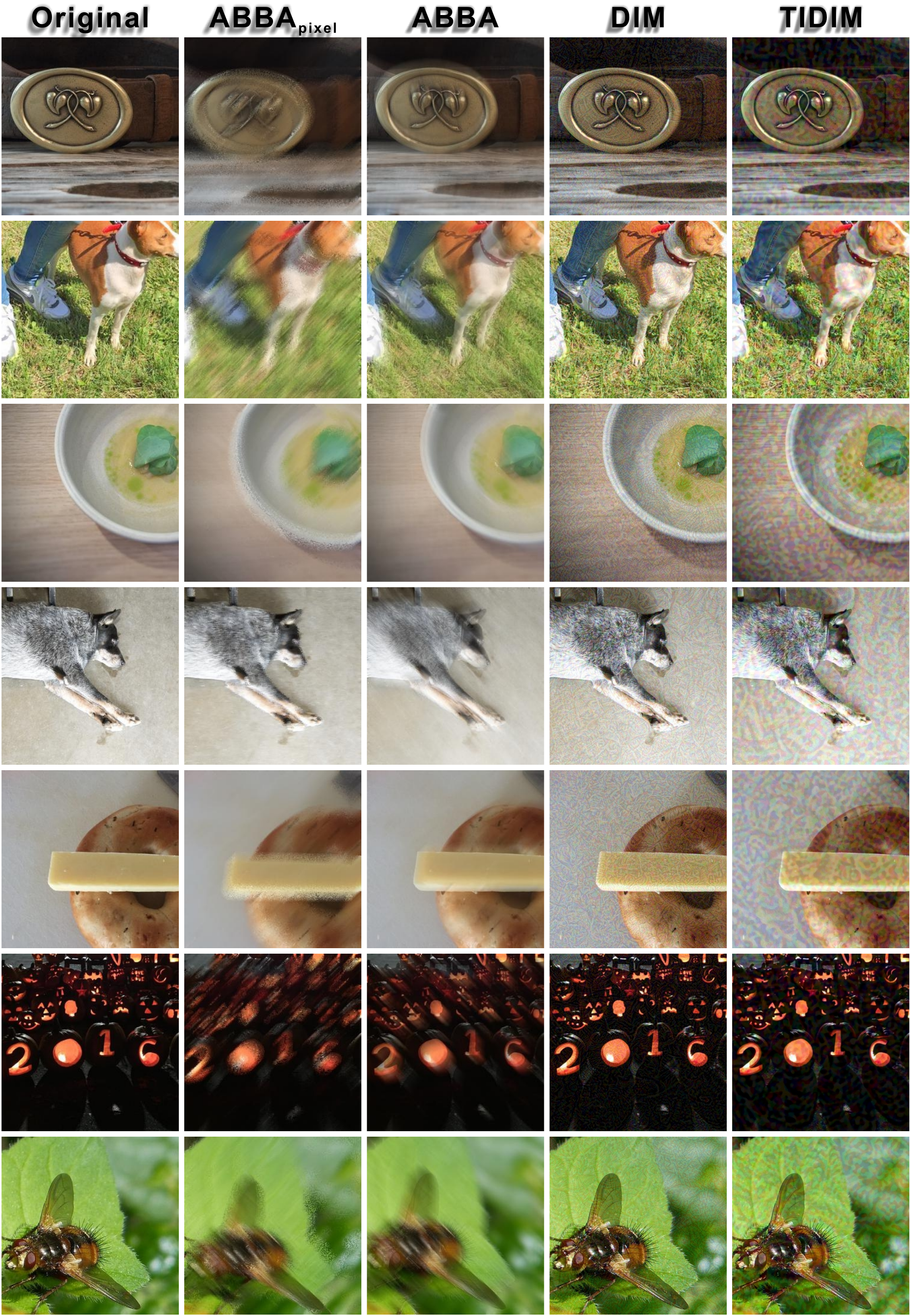}
\caption{Seven visualization results of AB$\kern-3pt$\textcolor{mygray}{B}A$_\text{pixel}$, AB$\kern-3pt$\textcolor{mygray}{B}A, DIM, and TIDIM. All adversarial examples fool the Inc-v3 model.}
\label{fig:cmp_vis}
\vspace{-0em}
\end{figure}

\begin{figure}[h]
\vspace{-1em}
\renewcommand\thefigure{X}
\centering
\includegraphics[width=1.0\columnwidth]{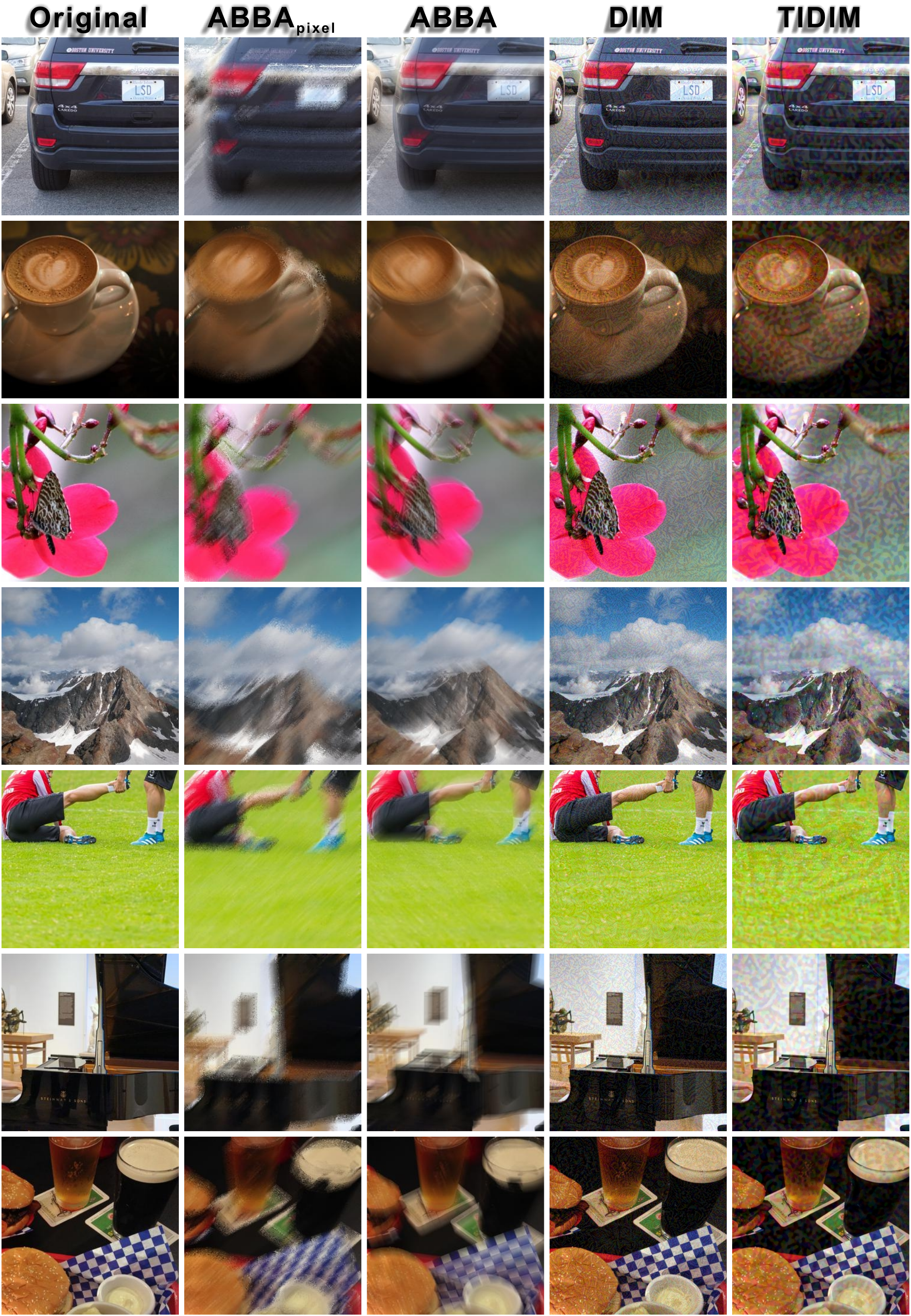}
\caption{Seven visualization results of AB$\kern-3pt$\textcolor{mygray}{B}A$_\text{pixel}$, AB$\kern-3pt$\textcolor{mygray}{B}A, DIM, and TIDIM. All adversarial examples fool the Inc-v3 model.}
\label{fig:cmp_vis1}
\vspace{-0em}
\end{figure}

We show several adversarial examples of AB$\kern-3pt$\textcolor{mygray}{B}A$_\text{pixel}$, AB$\kern-3pt$\textcolor{mygray}{B}A, DIM, and TIDIM in Fig.~\ref{fig:cmp_vis0}, Fig.~\ref{fig:cmp_vis} and Fig.~\ref{fig:cmp_vis1}. All examples can mislead the Inc-v3 model. 

Obviously, our method AB$\kern-3pt$\textcolor{mygray}{B}A can generate visually natural motion-blurred examples on various objects and these examples are very similar to real images captured by real-world cameras where the motion blur is caused by object or camera moving. In contrast, the adversarial examples of DIM and TIDIM have obvious unreal patterns. The noise-like pattern of DIM is drastically different from the natural noise usually caused by the camera sensor, \eg, Gaussian noise. The perturbation pattern of TIDIM is more perceptible than that of DIM, although TIDIM achieves much higher transferbility than DIM. Compared with AB$\kern-3pt$\textcolor{mygray}{B}A, our another method, \ie, AB$\kern-3pt$\textcolor{mygray}{B}A$_\text{pixel}$, breaks local pattern of the original input. However, AB$\kern-3pt$\textcolor{mygray}{B}A$_\text{pixel}$'s examples look more imperceptible than TIDIM's results. 
More comparison results are shown in Fig.~\ref{fig:cmp_vis} and Fig.~\ref{fig:cmp_vis1}.

\begin{figure}[t]
\renewcommand\thefigure{II}
\begin{center}
\centerline{\includegraphics[width=0.65\linewidth]{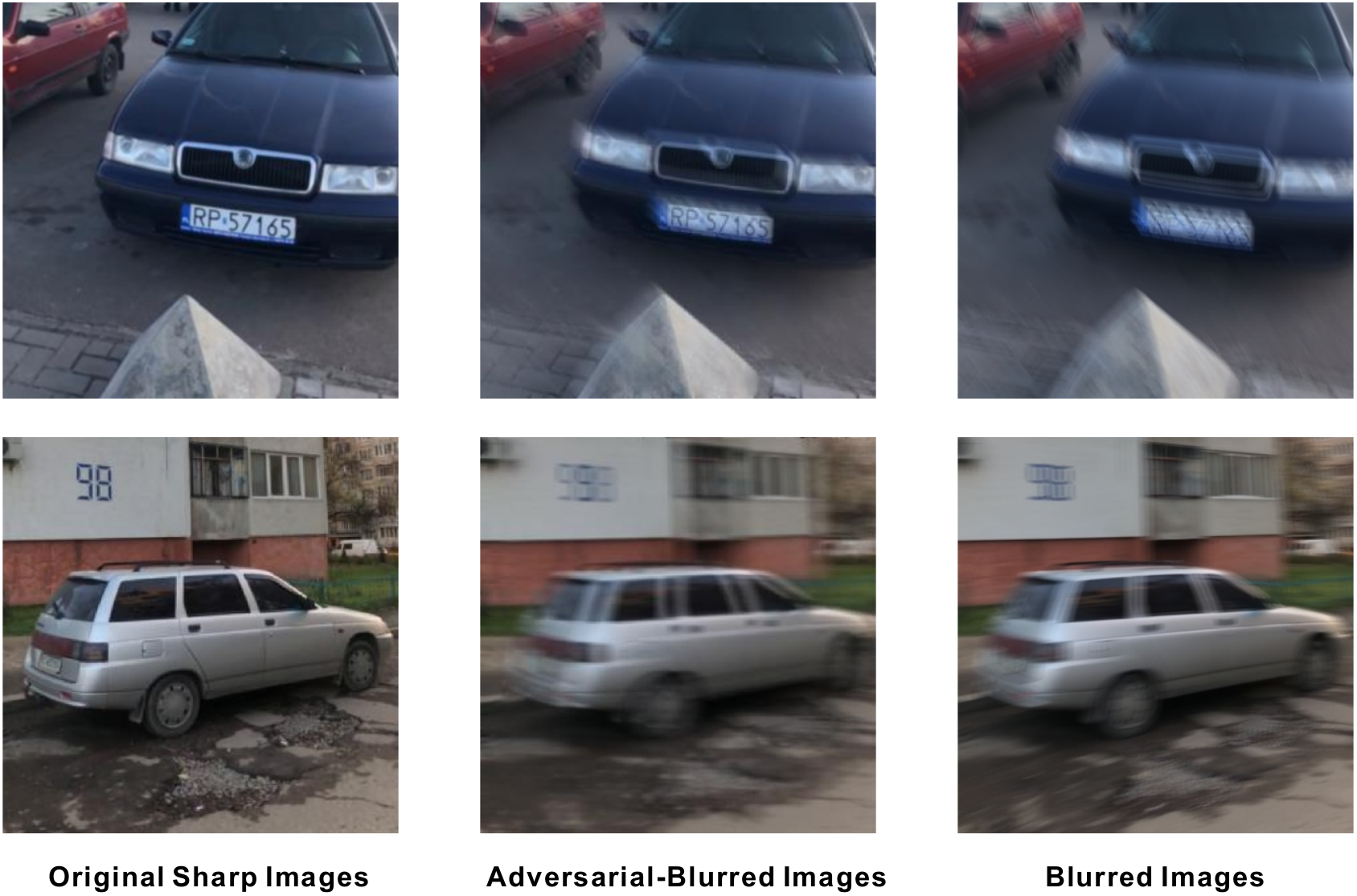}}
\caption{Comparison between adversarial-blurred images and blurred images for training deblurring models. 
}\vspace{-2em}
\label{fig:synvsadv}
\end{center}
\end{figure}

Besides above visualization results, we further conduct an experiment to compare our adversarial blur images with the blur images for training deblurring models~\cite{KupynCVPR2018}. Specifically, given a sharp image, \eg, the left sub-figures in Fig.~\ref{fig:synvsadv}, we use AB$\kern-3pt$\textcolor{mygray}{B}A to generate corresponding adversarial blur images and compare them with the blur images for training. Obviously, both blur looks realistic, which demonstrates the capability of AB$\kern-3pt$\textcolor{mygray}{B}A to generate visually natural blur images.

%--------------------------
%--------------------------

%-----------------------------------------------------------------------
\subsection{Attack Results of Inc-v3, Inc-v4, IncRes-v2, and Xception}

%--------------------------
\begin{table}[t]
	\caption{Adversarial comparison results on NeurIPS'17 adversarial competition dataset. 
	There is no available Xception model based on the author's implementations \cite{DongCVPR2019} of baselines, \ie, FGSM, MIFGSM, DIM, TIFGSM, TIFMIFGSM, and TIDIM. Hence, we leave these baselines' results empty for the Xception model. We highlight the top three results with \first{pink}, \second{yellow}, and \third{blue}, respectively. }%\vspace{-0em}
	\tiny
% 	\begin{center}
	\setlength\tabcolsep{1.5pt} % default value: 6pt
		\begin{tabular}{l|c|c|c|c|c|c|c|c|c|c|c|c|c|c|c|c}
			\hline
% 			\multirow{}{}{} 
			\rowcolor{tabgray2}~ & \multicolumn{4}{c|}{Attacking Results (Inc-v3)} & \multicolumn{4}{c|}{Attacking Results (Inc-v4)} & \multicolumn{4}{c|}{Attacking Results (IncRes-v2)} &
			\multicolumn{4}{c}{Attacking Results (Xception)} \\
			\cline{2-17}
			\rowcolor{tabgray2} & Inc-v3 & Inc-v4 & IncRes-v2 & Xception & Inc-v3 & Inc-v4 & IncRes-v2 & Xception & Inc-v3 & Inc-v4 & IncRes-v2 & Xception & Inc-v3 & Inc-v4 & IncRes-v2 & Xception\\
			\hline
			\hline
			{GaussBlur} & 34.7 & 22.7 & 18.4 & 26.1 & 14.2 & 26.7 & 10.9 & 17.2 & 12.1 & 11.8 & 20.1 & 13.8 & 16.1 & 15.7 & 11.9 & 32.5 \\
		    \cline{2-17}
			{GaussBlur}$_\text{obj}$ & 13.6 & 6.0 & 5.2 & 7.1 & 3.5 & 9.5 & 2.2 & 3.9 & 3.2 & 2.8 & 6.4 & 2.7 & 3.7 & 3.4 & 2.6 & 10.9 \\
			\cline{2-17}
			{GaussBlur}$_\text{bg}$ & 18.8 & 10.8 & 9.2 & 12.0 & 6.7 & 13.4 & 5.5 & 7.3 & 6.7 & 6.5 & 11.8 & 6.8 & 7.6 & 7.1 & 6.3 & 16.3 \\
		    \hline
			{DefocBlur} & 30.0 & 16.8 & 11.1 & 18.8 & 18.7 & 36.2 & 13.2 & 22.3 & 15.8 & 14.7 & 23.4 & 17.4 & 18.5 & 18.7 & 12.9 & 36.8\\
		    \cline{2-17}
			{DefocBlur}$_\text{obj}$ & 10.0 & 3.0 & 2.9 & 3.6 & 3.9 & 10.3 & 3.1 & 4.4 & 3.8 & 3.2 & 7.5 & 3.2 & 4.4 & 4.4 & 2.6 & 11.8 \\
			\cline{2-17}
			{DefocBlur}$_\text{bg}$ & 16.9 & 9.2 & 7.0 & 10.5 & 10.4 & 20.1 & 8.3 & 12.8 & 8.9 & 9.4 & 15.3 & 10.1 & 10.2 & 11.4 & 8.4 & 21.6 \\
			\hline
			{Interp}$_\text{blur}$ & 34.7 & 3.6 & 0.5 & 3.4 & 2.7 & 26.7 & 0.8 & 3.1 & 3.1 & 3.1 & 20.1 & 3.4 & 3.0 & 3.1 & 0.8 & 32.5 \\
			\hline
			\hline
			\rowcolor{tabgray1}{ AB$\kern-3pt$\textcolor{mygray}{B}A}$_\text{obj}$ & 21.0 & 4.9 & 4.2 & 7.0 & 11.6 & 28.9 & 9.7 & 11.5 & 11.2 & 11.9 & 29.0 & 12.7 & 9.1 & 9.6 & 7.7 & 30.2 \\
			\cline{2-17}
			\rowcolor{tabgray1}{ AB$\kern-3pt$\textcolor{mygray}{B}A}$_\text{bg}$ & 30.9 & 11.6 & 10.1 & 12.9 & 14.0 & 31.7 & 13.3 & 15.7 & 14.0 & 14.0 & 25.8 & 13.2 & 12.5 & 14.3 & 11.4 & 33.3 \\
			\cline{2-17}
			\rowcolor{tabgray1}{ AB$\kern-3pt$\textcolor{mygray}{B}A}$_\text{image}$ & 62.4 & 29.8 & 28.8 & 34.1 & 32.0 & 66.7 & 28.8 & 36.2 & 33.0 & 30.7 & 63.4 & 37.0 & 28.9 & 28.4 & 26.1 & 66.7 \\
			\cline{2-17}
			\rowcolor{tabgray1}{ AB$\kern-3pt$\textcolor{mygray}{B}A}$_\text{pixel}$ & 89.2 & \cellcolor{top3} 65.5 & \cellcolor{top3} 65.8 & \cellcolor{top2} 71.2 & \cellcolor{top3} 77.7 & 88.1 & \cellcolor{top3} 71.3 & \cellcolor{top2} 76.0 & \cellcolor{top3} 81.8 & \cellcolor{top3} 78.3 & 92.0 & \cellcolor{top2} 80.6 & 74.0 & 67.5 & 66.8 & 86.2\\
			\cline{2-17}
			\rowcolor{tabgray1}{ AB$\kern-3pt$\textcolor{mygray}{B}A} & 65.6 & 31.2 & 29.7 & 33.5 & 39.5 & 74.9 & 37.3 & 43.2 & 38.4 & 38.6 & 71.6 & 44.2 & 32.3 & 35.2 & 35.9 & 73.1 \\
			\hline
			\hline
			{Interp}$_\text{noise}$ & 95.8 & 20.5 & 15.6 & 22.9 & 5.2 & 92.6 & 1.6 & 6.0 & 6.7 & 6.0 & 91.8 & 8.3 & 3.5 & 2.3 & 0.4 & 93.4 \\
		    \hline
			{FGSM} & 79.6 & 35.9 & 30.6 & 42.1 & 43.1 & 72.6 & 32.5 & 45.2 & 44.3 & 36.1 & 64.3 & 45.4 & \cellcolor{tabgray3} & \cellcolor{tabgray3} & \cellcolor{tabgray3} &  \cellcolor{tabgray3}  \\
			\hline
			{MIFGSM} & \cellcolor{top3} 97.8 & 47.1 & 46.4 & 47.7 & 67.1 & \cellcolor{top1} 98.8 & 54.3 & 58.5 & 74.8 & 64.8 & \cellcolor{top1} 100.0 & 61.7 & \cellcolor{tabgray3} & \cellcolor{tabgray3} &  \cellcolor{tabgray3}  &  \cellcolor{tabgray3}  \\
			\hline
		    {DIM} &  \cellcolor{top2} 98.3 & \cellcolor{top2} 73.8 & \cellcolor{top2} 67.8 & \cellcolor{top1} 71.6 & \cellcolor{top1} 81.8 & \cellcolor{top3} 98.2 & \cellcolor{top1} 74.2 & \cellcolor{top1} 79.1 & \cellcolor{top2} 86.1 & \cellcolor{top2} 83.5 & \cellcolor{top2} 99.1 & \cellcolor{top1} 80.8 &  \cellcolor{tabgray3}  & \cellcolor{tabgray3}  & \cellcolor{tabgray3} &  \cellcolor{tabgray3} \\
			\hline
			{TIFGSM} & 75.4 & 37.3 & 32.1 & 38.6 & 45.3 & 68.1 & 33.7 & 39.4 & 49.7 & 41.5 & 63.7 & 44.0 & \cellcolor{tabgray3} & \cellcolor{tabgray3} & \cellcolor{tabgray3} &  \cellcolor{tabgray3}  \\
			\hline
			{TIMIFGSM} & 97.9 & 52.4 & 47.9 & 44.6 & 68.6 & \cellcolor{top1} 98.8 & 55.3 & 50.8 & 76.1 & 69.5 & \cellcolor{top1} 100.0 & 59.9 & \cellcolor{tabgray3} & \cellcolor{tabgray3}  &  \cellcolor{tabgray3}  &  \cellcolor{tabgray3}  \\
			\hline
			{TIDIM} & \cellcolor{top1} 98.5 &  \cellcolor{top1} 75.2 & \cellcolor{top1} 69.2 & \cellcolor{top3} 61.3 & \cellcolor{top2} 80.7 & \cellcolor{top2} 98.7 & \cellcolor{top2} 73.2 & \cellcolor{top3} 65.5 & \cellcolor{top1} 86.4 & \cellcolor{top1} 85.5 & \cellcolor{top3} 98.8& \cellcolor{top3} 71.0 &  \cellcolor{tabgray3}  &  \cellcolor{tabgray3}  &  \cellcolor{tabgray3}  &  \cellcolor{tabgray3}  \\
			\hline
		\end{tabular}
% 	\end{center}
	\label{tab:attack_results}
	\vspace{-1em}
\end{table}
Besides the attack results of Inc-v3 reported in our main paper, we further show the results of Inc-v4, IncRes-v2, and Xception in Table~\ref{tab:attack_results}. Note that, there is no available Xception model based on the authors' implementations \cite{DongCVPR2019} of FGSM, MIFGSM, DIM, TIFGSM, TIFMIFGSM, and TIDIM. Hence, we leave these baselines' results empty for the Xception model. Similar to the results of Inc-v3, for the transferability results, our method, \ie, AB$\kern-3pt$\textcolor{mygray}{B}A$_\text{pixel}$, achieves slightly lower success rate than the state-of-the-art additive-perturbation-based attacks, \ie, DIM and TIDIM, when attacking Inc-v3, Inc-v4, and IncRes-v2, and obtains higher success rate than TIDIM when attacking the Xception model. For the whitebox attacks, TIMIFGSM and MIFGSM usually achieve the highest success rate.

%-----------------------------------------------------------------------
\subsection{Hyper-parameter Analysis and Ablation Study}

%------------------------
% \begin{wrapfigure}{r}{0.49\columnwidth}
\begin{figure}[t]
\begin{center}
\renewcommand\thefigure{III}
\centerline{\includegraphics[width=0.70\columnwidth]{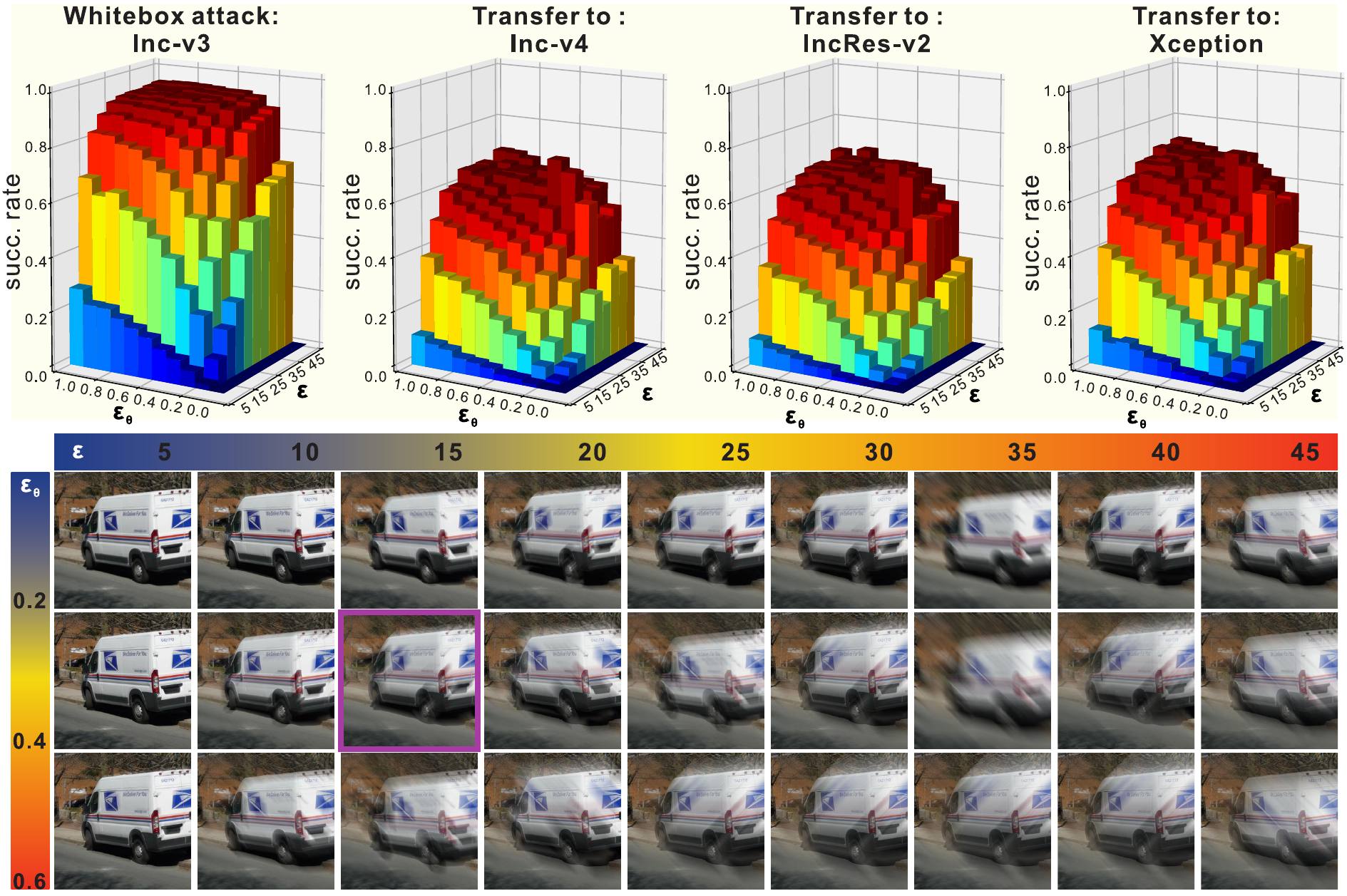}}
\caption{Up: shows the success rate of AB$\kern-3pt$\textcolor{mygray}{B}A \wrt the variation of both $\epsilon$ and $\epsilon_\theta$ in Eq.~(5) of our main paper where $\epsilon$ is within $[5,50]$ with step size $5$ and $\epsilon_\theta$ is in $[0,1]$ with step size $0.1$. Down: shows an example of .}
\label{fig:succ_rate2ksz_trans}
\vspace{-1em}
\end{center}
\end{figure}
% \end{wrapfigure}
%------------------------

{\bf Effect of $\epsilon$ and $\epsilon_\theta$.}
We calculate the success rate of our method with different $\epsilon$ and $\epsilon_\theta$ in the Eq.~(5) of our main paper, respectively. Specifically, we try $\epsilon$ with the range $[5,N]$ where $N=51$ and $\epsilon_\theta$ in $[0,1]$. 
As shown in Fig.~\ref{fig:succ_rate2ksz_trans} (a), the success rates become gradually higher with the increase of $\epsilon$ and $\epsilon_\theta$. The highest success rates are $94.8\%$, $68.5\%$ $68.4\%$, and $72.1\%$ on Inc-v3, Inc-v4, IncRes-v2, and Xception, respectively.
%
% when we set $\epsilon=1.0$ and $\epsilon_\theta=50.0$. 
% The success rate becomes zero when $\epsilon=0.0$ and $\epsilon_\theta=5.0$. 
We also visualize adversarial examples of an image that has been successfully attacked on all $\epsilon>0$ and $\epsilon_\theta>0$. Obviously, as $\epsilon$ and $\epsilon_\theta$ increase, the visual effects of adversarial examples gradually become worse and the perturbations are more easily perceived. 
According to numerous attacking on different images, we choose the $\epsilon=15.0$ and $\epsilon_\theta=0.4$ to balance the success rate and visual effects when comparing with baselines on transferability in Sec.~3.2 in our main paper.

%------------------------
%\begin{wrapfigure}{r}{0.5\columnwidth}
\begin{figure}[t]
\centering
\renewcommand\thefigure{IV}
\includegraphics[width=0.65\columnwidth]{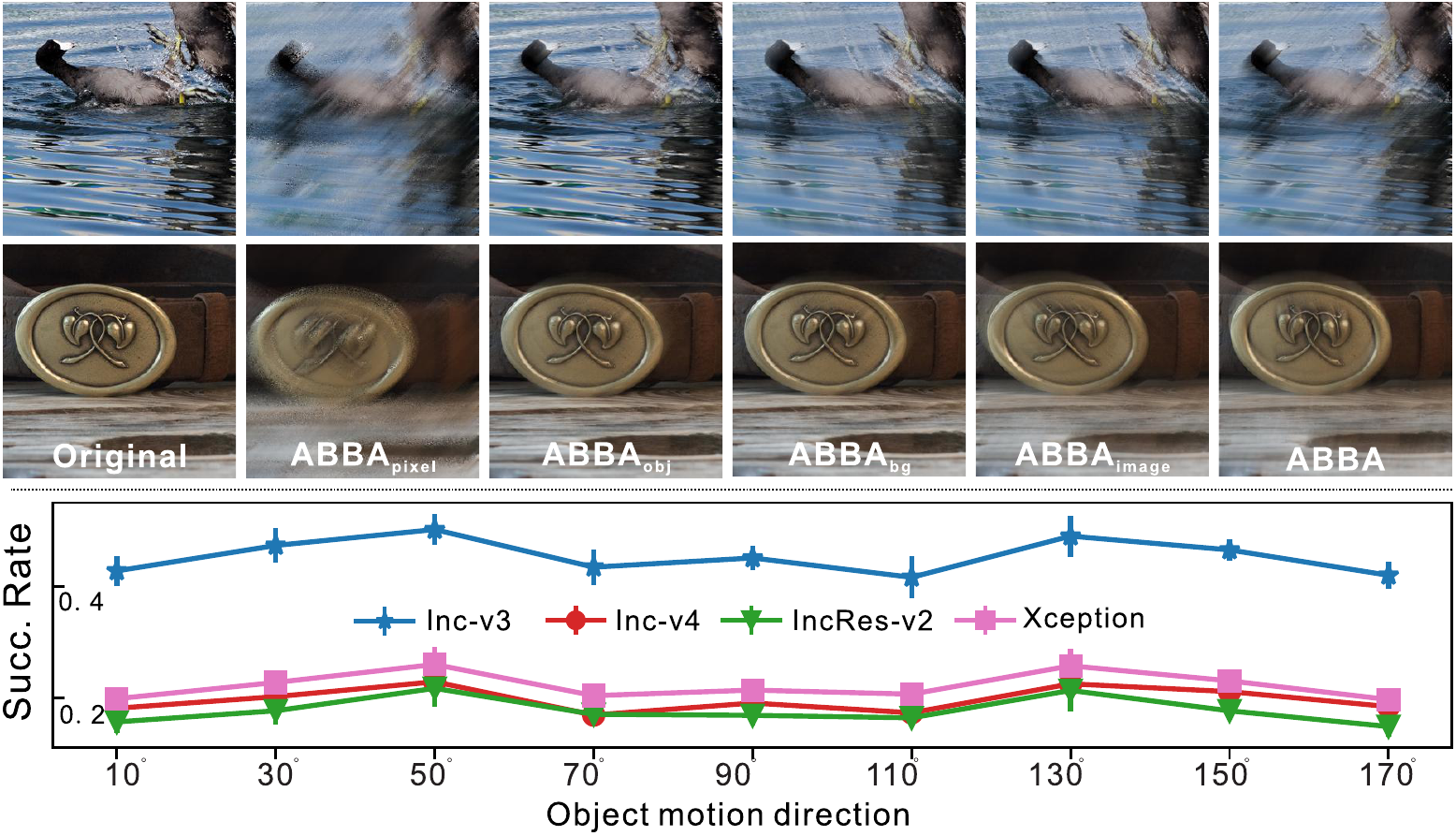}
\caption{Up: two examples of AB$\kern-3pt$\textcolor{mygray}{B}A$_\text{pixel}$,  AB$\kern-3pt$\textcolor{mygray}{B}A$_\text{obj}$, AB$\kern-3pt$\textcolor{mygray}{B}A$_\text{bg}$, AB$\kern-3pt$\textcolor{mygray}{B}A$_\text{image}$, and AB$\kern-3pt$\textcolor{mygray}{B}A. Bottom: Success rates of our method with respect to the object motion directions.}
\label{fig:vis_ablation}
\vspace{-1.0em}
\end{figure}
%\end{wrapfigure}
%------------------------
%

\noindent{\bf Effect of motion directions.} we fix $\epsilon_\theta=0.4$ and $\epsilon=15.0$ and tune the motion direction of object and background by setting different x-axis and y-axis translations. For each object motion direction, we calculate the mean and standard variation of the success rates on different background moving directions. As shown in Fig.~\ref{fig:vis_ablation}~(B), the success rate increases as the object motion direction becomes larger in $[10^\circ,50^\circ]$ while decreasing as the direction is smaller in $[50^\circ,70^\circ]$. The success rate variation has symmetrical trend in the range of $[90^\circ,170^\circ]$. Such results are mostly caused by the $L\infty$ used for constraining the translation. The motion direction is directly related to the translation and the success rate reaches the highest value around $45^\circ$.
% For example, with the motion direction of $45^\circ$ and the constraints of x-axis and y-axis translation $\epsilon_\theta$, the allowed translation of object is $\sqrt{2}\epsilon_\theta$ that is the maximum allowed translation among different motion directions. 
% As a result, the success rate reach the highest value around $45^\circ$.

\noindent{\bf Effect of blurred regions and importance of adaptive translations.}
As reported in Tab.~1 in our main paper and cases shown in Fig.~\ref{fig:vis_ablation}~(U), AB$\kern-3pt$\textcolor{mygray}{B}A$_\text{pixel}$ achieves the highest attack success rate and transferability among all variants, which, however, changes the original image obviously and looks unnatural. AB$\kern-3pt$\textcolor{mygray}{B}A$_\text{obj}$ and AB$\kern-3pt$\textcolor{mygray}{B}A$_\text{bg}$ have the worst success rate on all models although they tend to generate visually natural motion blur. AB$\kern-3pt$\textcolor{mygray}{B}A$_\text{image}$ and AB$\kern-3pt$\textcolor{mygray}{B}A make good balance between the attack success rate and visual effects. In particular, AB$\kern-3pt$\textcolor{mygray}{B}A that jointly tunes the object and background translations can obtain much better transferability across normal trained and defense-based models. 
Note that, when compared with the results using fixed motion directions in Fig.~\ref{fig:vis_ablation}~(B), AB$\kern-3pt$\textcolor{mygray}{B}A obtains the highest success rate among all motion direction, further demonstrating usefulness of adaptive translations.

%-----------------------------------------------------------------------
\subsection{Interpretable Explanation of the Transferability}
In the following, we explore the difference between AB$\kern-3pt$\textcolor{mygray}{B}A, FGSM, and MIFGSM on the transferability. Note that, we implement FGSM and MIFGSM on the same platform (\ie, pytorch with foolbox~2.3.0) with AB$\kern-3pt$\textcolor{mygray}{B}A for fair comparison.
We modify the method in \cite{FongICCV2017} that generates an interpretable map for a classification model $\mathrm{f}(\cdot)$ with a given perturbation.
Then, we observe that the transferability of an adversarial example generated by an attack correlates with the consistency of interpretable maps of different models. 
Specifically, given an adversarial example $\mathbf{X}^{\text{adv}}$ generated by an attack and the original image $\mathbf{X}^{\text{real}}$, we can calculate an interpretable map $\mathbf{M}^{\mathrm{f}}$ for $\mathrm{f}(\cdot)$ by optimizing:
%
%---------------------------
\begin{align}\label{eq:interp_map}
\argmin_{\mathbf{M}^{\mathrm{f}}} ~~\mathrm{f}_y(\mathbf{M}^{\mathrm{f}} \odot \mathbf{X}^\text{adv} + (1-\mathbf{M}^{\mathrm{f}}) \odot \mathbf{X}^\text{real})
+ \lambda_{1} \| \mathbf{M}^{\mathrm{f}} \|_{1} + \lambda_{2}\mathrm{TV} (\mathbf{M}^{\mathrm{f}}) 
\end{align}
%---------------------------
%
where $\mathrm{f}_y(\cdot)$ denotes the score at label $y$ that is the ground truth label of $\mathbf{X}^{\text{real}}$ and $\mathrm{TV}(\cdot)$ is the total-variation norm. Intuitively, optimizing Eq.~(\ref{eq:interp_map}) is to find the region that causes misclassification. 
%
%---------------------------
% \begin{wrapfigure}{r}{0.48\columnwidth}
% \vspace{-1em}
\begin{figure}[t]
\renewcommand\thefigure{V}
\centering
\includegraphics[width=0.65\columnwidth]{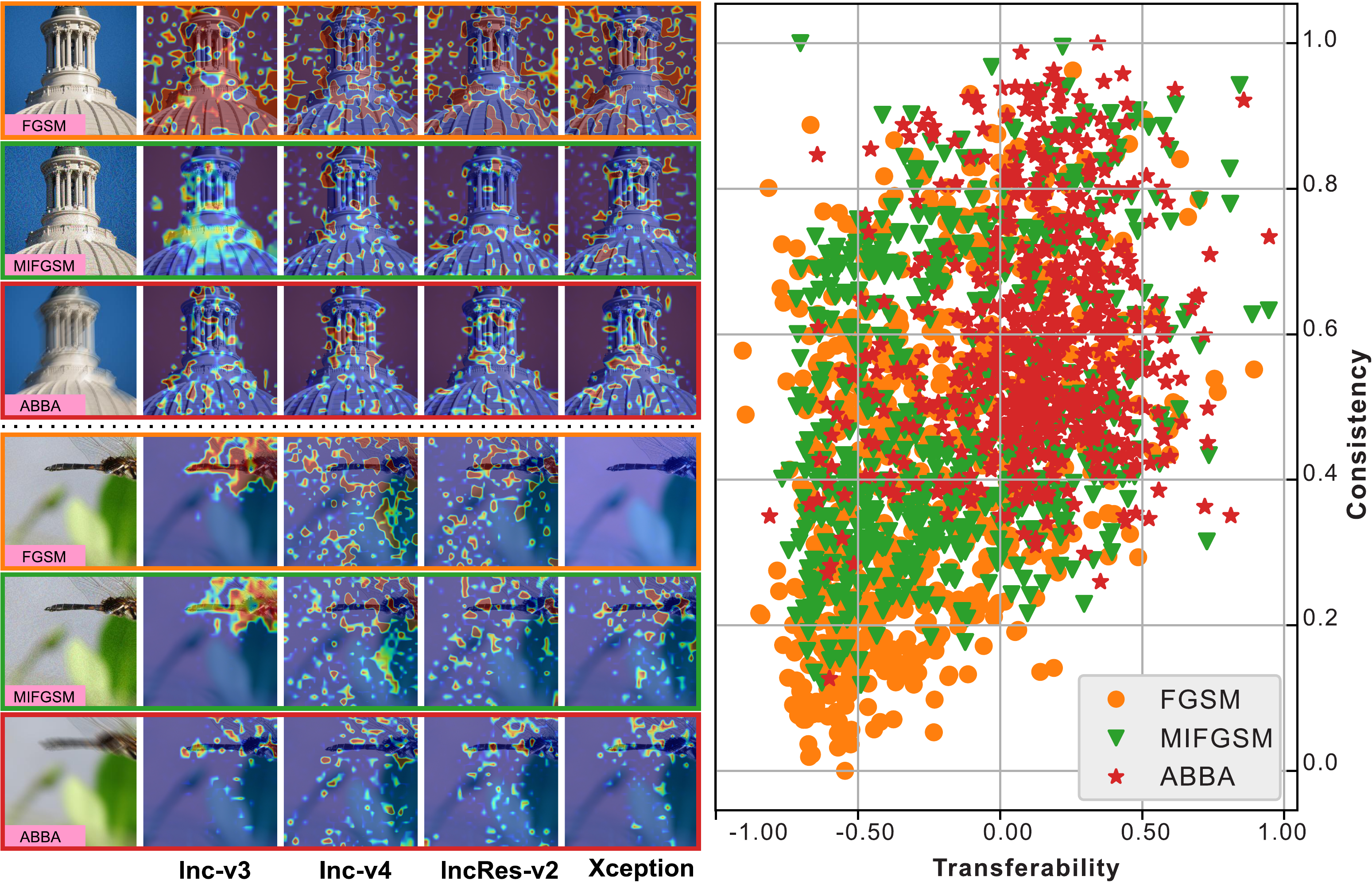}
\caption{Left: the interpretable maps of six adversarial examples generated by FGSM, MIFGSM, and AB$\kern-3pt$\textcolor{mygray}{B}A, respectively, with four models. Right: the transferability \& consistency distributions of adversarial examples generated by the three attacks.}
\label{fig:interp_pert}
\end{figure}
% \vspace{-1em}
% \end{wrapfigure}
%---------------------------
%
We optimize Eq.~(\ref{eq:interp_map}) via gradient decent in 150 iterations and fix $\lambda_{1}=0.05$ and $\lambda_{2}=0.2$. We can calculate four interpretable maps for each adversarial example based on four models, \ie, Inc-v3, Inc-v4, IncRes-v2, and Xception, as shown in Fig.~\ref{fig:interp_pert}(L). We observe that the interpretable maps of our method have similar distributions across the four models while the maps of FGSM and MIFGSM do not exhibit this phenomenon. To further validate this observation, we calculate the standard variation across the four maps at each pixel and get a value by mean pooling. We normalize the value and regard it as the consistency measure for the four maps. As shown in Fig.~\ref{fig:interp_pert}(R), the consistency of adversarial examples of our method is generally higher than that of FGSM and MIFGSM.
We further study the transferability of an adversarial example across models. Given an adversarial example from Inc-v3 and a model $\mathrm{f}(\cdot)$, we calculate a score to measure the transferability under this model: $t^{\mathrm{f}}=\mathrm{f}_c(\mathbf{X}^{\text{adv}})-\mathrm{f}_y(\mathbf{X}^{\text{adv}})$  where $c\neq y$ is the label having maximum score among non-ground-truth labels.
If $t^{\mathrm{f}}>0$ means the adversarial example fool $\mathrm{f}(\cdot)$ successfully, and vice versa. As shown in Fig.~\ref{fig:interp_pert}(R), the transferability of adversarial examples of our method is generally higher than that of FGSM and MIFGSM.
%
% Based on the observations in Fig.~\ref{fig:interp_pert}(R), we conjecture that higher consistency across models leads to higher transferability. This is reasonable since the high consistency of interpretable maps means the adversarial example has the same effect to different models and the optimized perturbation on the whitebox model can be easily transferred to other models. 

%----------------------------------------------------------------------
\subsection{Attack results of STN-based model}
\begin{figure}[t]
\renewcommand\thefigure{VIII}
\centering
\includegraphics[width=0.65\columnwidth]{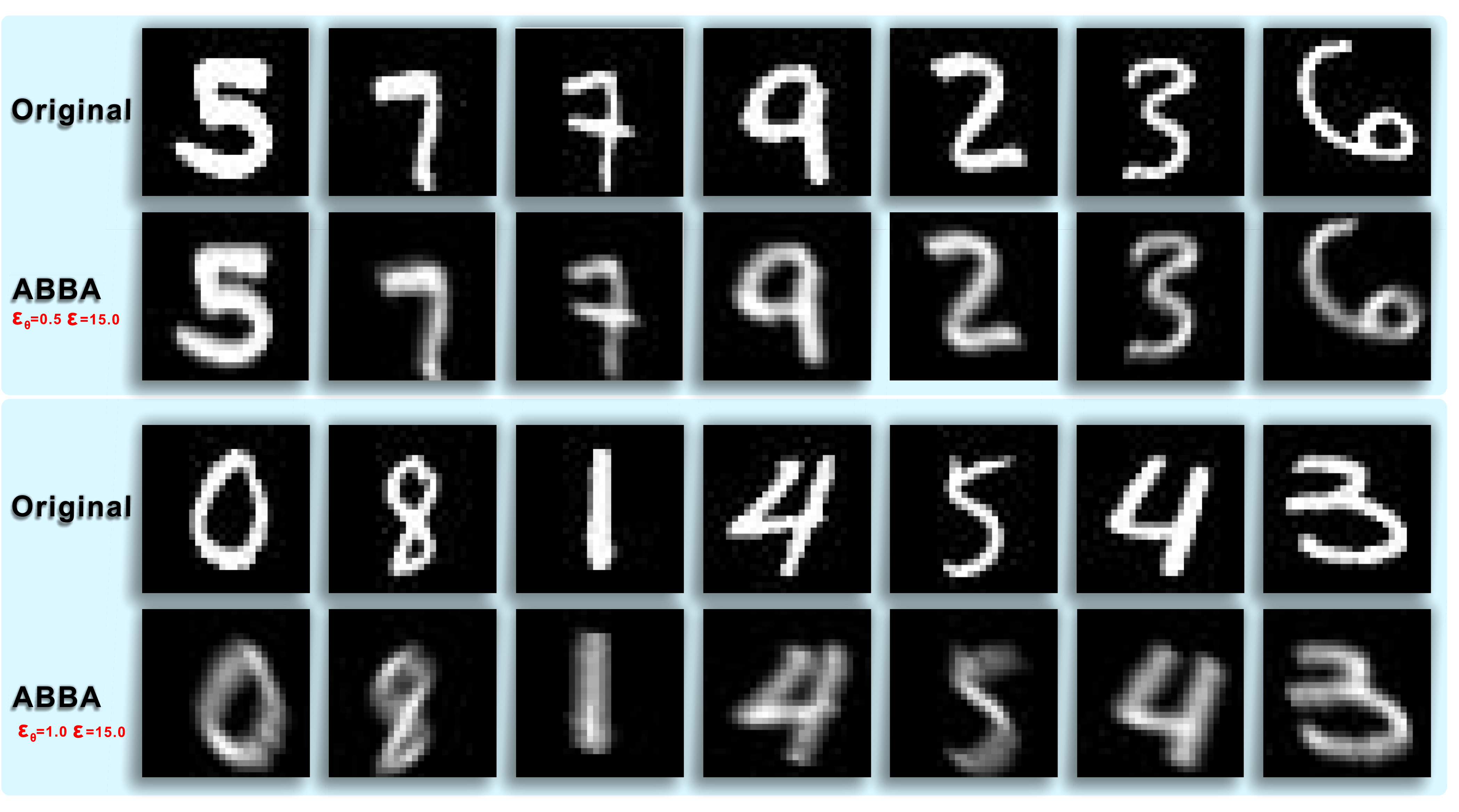}
\caption{Visualization examples of AB$\kern-3pt$\textcolor{mygray}{B}A for attacking STN-based CNN with two group hyper-parameters, \ie, $\epsilon_\theta=0.5,\epsilon=15.0$ and $\epsilon_\theta=1.0,\epsilon=15.0$.}
\label{fig:supp_stn_vis}
\vspace{-0em}
\end{figure}

As introduced in Sec.~2.3 in the main paper, we employ the spatial transformer network~(STN) to tune the translation parameters of the object and background, and it may post a question if our method could also be useful in attacking STN-based CNN models.
%
%
%---------------------------
\begin{wrapfigure}{r}{0.48\columnwidth}
\vspace{-1em}
\renewcommand\thefigure{VI}
\centering
\includegraphics[width=0.40\columnwidth]{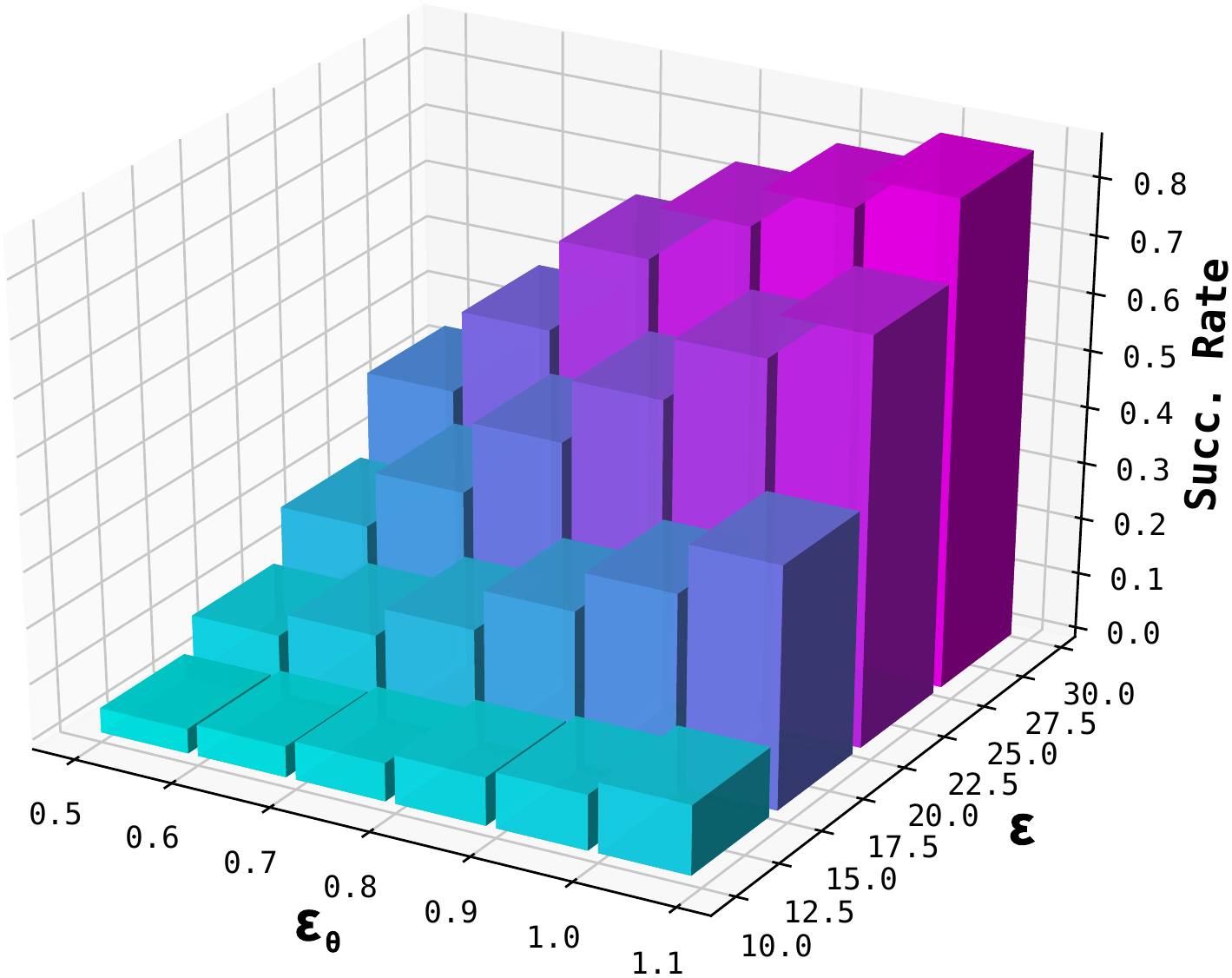}
\caption{Success rate of AB$\kern-3pt$\textcolor{mygray}{B}A w.r.t. different $\epsilon_\theta$ and $\epsilon$ that are maximum translations and maximum number of valid kernel elements.}
\label{fig:supp_stn_succrate}
% \end{figure}
\vspace{-1em}
\end{wrapfigure}
%---------------------------
%
%
Towards more comprehensive evaluation and answering this question, we further conduct an experiment to attack the STN-based CNN implemented by \cite{stn_url} on the MNIST dataset. \textit{First}, we use our AB$\kern-3pt$\textcolor{mygray}{B}A to attack the STN-based CNN with different hyper-parameters, \ie, the maximum translations~($\epsilon_\theta$) and the maximum number of valid kernel elements~($\epsilon_\theta$). As shown in Fig.~\ref{fig:supp_stn_succrate}, the success rate of our method gradually increases as the $\epsilon_\theta$ and $\epsilon$ become larger, which demonstrates the effectiveness of our method. \textit{Second}, we show several examples of our attack results in Fig.~\ref{fig:supp_stn_vis} with two groups of hyper-parameters, \ie, $\epsilon_\theta=0.5,\epsilon=15.0$ and $\epsilon_\theta=1.0,\epsilon=15.0$. As shown in Fig.~\ref{fig:supp_stn_vis}, with the small $\epsilon_\theta=0.5$, our method can generate slightly blurred handwritten digits that look naturally but fool the STN-based CNN. When using larger $\epsilon_\theta=1.0$, the adversarial images become more blurred but still natural. 

%----------------------------------------------------------------------
\subsection{Benefits to Blur-Robustness Enhancement}

We conduct an experiment that trains the IncResv2 on the clean imagenet containing 1.1 million images and a modified imagenet with 1.3 million images including 0.2 million AB$\kern-3pt$\textcolor{mygray}{B}A-blurred images, and evaluate the accuracy on the motion-blur subsets of ImageNetC \cite{hendrycks2019robustness}. We see that the Top-1 error of IncResv2 decreases from 73.0\% to 53.2\% with our blurred images, which strongly demonstrates the impact of AB$\kern-3pt$\textcolor{mygray}{B}A to enhance the blur-robustness of DNNs.

\subsection{More results of  AB$\kern-3pt$\textcolor{mygray}{B}A$_\text{physical}$}

\begin{figure}[t]
\renewcommand\thefigure{VII}
\centering
\includegraphics[width=0.65\columnwidth]{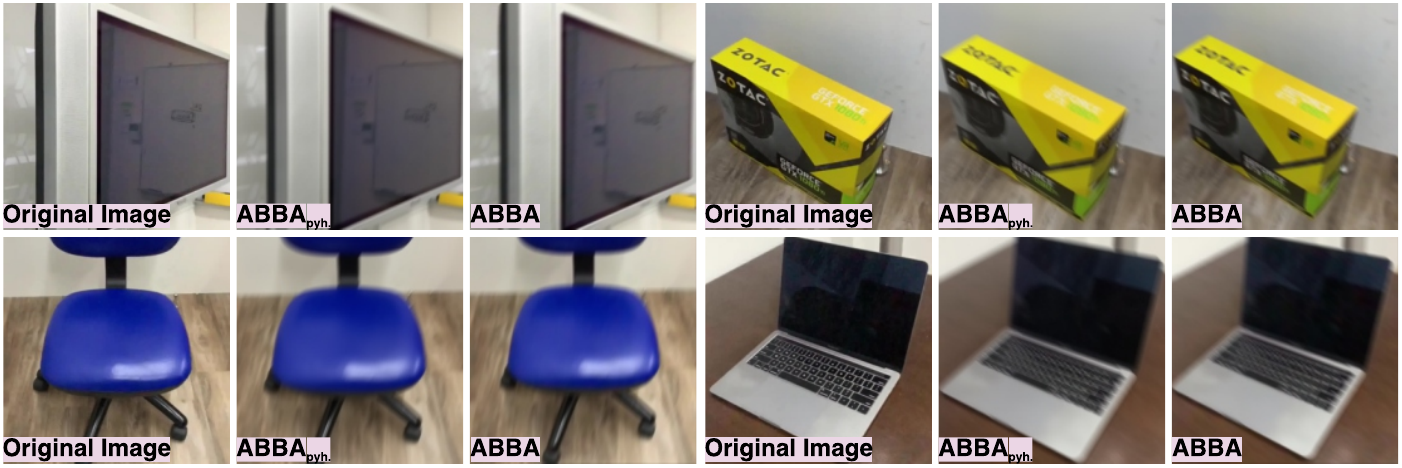}
\caption{Comparing the visualization examples of AB$\kern-3pt$\textcolor{mygray}{B}A$_\text{physical}$ with those of AB$\kern-3pt$\textcolor{mygray}{B}A.}
\label{fig:abbasupp_sys_vs_real}
\vspace{-0em}
\end{figure}

In Figure~\ref{fig:abbasupp_sys_vs_real}, we present four more examples generated by AB$\kern-3pt$\textcolor{mygray}{B}A$_\text{physical}$ that captures real-world images according the camera moving parameters estimated by AB$\kern-3pt$\textcolor{mygray}{B}A. 
We try our best to compare synthesis images from AB$\kern-3pt$\textcolor{mygray}{B}A with the physical images from AB$\kern-3pt$\textcolor{mygray}{B}A$_\text{physical}$, demonstrating the motion blurred images generated by our method could be really found in the real world.

%--------------------------------------
\begin{table}[t]
\renewcommand\thetable{III}
\centering
\tiny
% \footnotesize
\setlength\tabcolsep{6pt} % default value: 6pt
    % \vspace{-0.5em}
    \caption{Succ. rates of AB$\kern-3pt$\textcolor{mygray}{B}A and NormalBlur before (Adv. from Inc-v3) and after deblurring via original and re-trained DeblurGANv2s.}
    % \vspace{-0.5em}
    % 	\tiny
% 	\setlength\tabcolsep{1.5pt} % default value: 6pt
		\begin{tabular}{l|c|c|c|c|c|c|c|c|c}
			\hline
			\rowcolor{tabgray2}~ & \multicolumn{3}{c|}{ Adv. from Inc-v3} & \multicolumn{3}{c|}{DeblurGANv2} & \multicolumn{3}{c}{Re-DeblurGANv2} \\
			\cline{2-10}
			\rowcolor{tabgray2} & Inc-v3 & Inc-v4 & IncRes-v2 & Inc-v3 & Inc-v4 & IncRes-v2 & Inc-v3 & Inc-v4 & IncRes-v2 \\
			\hline
			\hline
			AB$\kern-3pt$\textcolor{mygray}{B}A & 65.3 & 31.1 & 30.0 & 31.4 & 24.2 & 18.4 & 22.9 & 22.5 & 16.8 \\
		    \cline{2-10}
			NormalBlur & 36.4 & 20.8 & 18.5 & 15.2 & 10.0 & 4.7 & 26.4 & 18.1 & 13.7 \\
			\hline
		\end{tabular}
		\label{tab:comp}
% 	\vspace{-2em}
\end{table}
% --------------------------------------
%

\subsection{Defense Results of Re-trained DeblurGANv2 on AB$\kern-3pt$\textcolor{mygray}{B}A and NormalBlur}
NormalBlur generates motion-blurred image by optimizing Eq.~(5) while fixing all kernel elements as $\frac{1}{N}$, which is equivalent to averaging neighbouring video frames where object and background move uniformly.
In contrast, AB$\kern-3pt$\textcolor{mygray}{B}A effectively tunes kernel elements to fool DNNs.
Actually, the intention of Sec 3.5 is to study the effectiveness of existing deblurring method (\ie, the `already-deployed' deblurring modules)  in defending the attack of AB$\kern-3pt$\textcolor{mygray}{B}A with the tunable kernels. More detailed response:
\ding{182} {\bf NormalBlur} utilizes Eq.~(5) to generate motion blur and {\bf has considered  background motion} via optimizing $\theta_\text{b}$.
\ding{183} 
In practice, our assumption is that {\bf we cannot get real motion information in the scene} and there is only one given static image. Hence, our attack is conducted under this assumption, \ie, the object and background move uniformly (\ie, at fixed speed) in a short time, which is a common phenomenon in the real world (\eg, walking). Our attack could be easily extended to other cases where more motion information is available (\eg, video).
\ding{184} 
With the DeblurGANv2 trained on normal motion blur dataset (\eg, GOPRO [22]), the decrease in succ. rate before and after deblurring in Table~\ref{tab:comp} have shown that the NormBlur can be defended more easily than AB$\kern-3pt$\textcolor{mygray}{B}A, which demonstrates AB$\kern-3pt$\textcolor{mygray}{B}A's tunable kernels facilitate achieving high attack success rate and anti-deblurring capability.
{\bf As suggested by the reviewer, when we further retrained the DeblurGANv2 with blurred images from AB$\kern-3pt$\textcolor{mygray}{B}A.} AB$\kern-3pt$\textcolor{mygray}{B}A can be defended more easily,
which further indicates a promising direction of combining AB$\kern-3pt$\textcolor{mygray}{B}A and the existing deblurring method for effective defense.

%---------------------------
%---------------------------

%---------------------------
%---------------------------

%---------------------------
%---------------------------

%---------------------------
%---------------------------

\small

\bibliographystyle{unsrt}
\bibliography{neurips_2020}

\end{document}